\documentclass[11pt]{article}

\usepackage[final]{acl}

\usepackage{times}
\usepackage{latexsym}
\usepackage[T1]{fontenc}
\usepackage[utf8]{inputenc}
\usepackage{microtype}
\usepackage{marvosym}   % provides \Letter (envelope icon)

\usepackage{graphicx}
\usepackage{booktabs}
\usepackage{multirow}
\usepackage{algorithm}
\usepackage{algorithmic}
\usepackage{amsmath, amssymb}
\usepackage{xcolor}
\usepackage{colortbl}
\usepackage[capitalize,noabbrev]{cleveref}

\title{Enjoy Your Talk: A Human-Centered Benchmark for Multi-Turn Dialogue with Decoupled User Simulation, Target Modeling, and Judging}

\author{
\bfseries Jinglan Gong$^{1,2}$\textsuperscript{*}\hspace{0.3cm}
Jiefan Lu$^{1}$\textsuperscript{*}\hspace{0.3cm}
Hewei Guo$^{1}$\textsuperscript{*}\hspace{0.3cm}
Kehan Li$^{1,3}$\hspace{0.3cm}
Zhiyuan Han$^{1,2}$\hspace{0.3cm}
Jihang Jiang$^{2}$\\
\bfseries Wenwen Tong$^{1}$\textsuperscript{\Letter}\hspace{0.3cm}
Lewei Lu$^{1}$\\
\normalfont
$^1$SenseTime Research\hspace{0.4cm}
$^2$University of Science and Technology of China\hspace{0.4cm}
$^3$Tsinghua University\\
}

\begin{document}
\maketitle

\begin{NoHyper}
\renewcommand{\thefootnote}{}\footnotetext{\textsuperscript{*}\,Equal contribution.\hspace{0.3cm}\textsuperscript{\Letter}\,Corresponding authors.}
\end{NoHyper}
\renewcommand{\thefootnote}{\arabic{footnote}}

%-------------------------------------------------------------------------
\begin{abstract}
Evaluating large language models (LLMs) as multi-turn conversational
partners requires probing capabilities that single-turn benchmarks miss:
persona consistency, evolving intent tracking, emotional dynamics, and
goal completion across many turns. We introduce \textbf{EYT-Bench}, a
human-centered benchmark whose evaluation protocol is built around a
\emph{decoupled three-party} design: a persona-grounded user
simulator, a target model evaluated on both intent perception and response generation, and an independent, configurable ensemble of LLM judges. Across $3{,}400$ dialogues with $17$ target models, EYT-Bench reveals four findings
that previous benchmarks miss: \textbf{(i)} state-of-the-art \emph{closed} and
\emph{open-source} models are statistically indistinguishable on
subjective dimensions, but separate by up to $9{\times}$ on objective
intent-tracking; \textbf{(ii)} reasoning is a phase
transition for objective tracking on long-context personas but is essentially
flat on subjective scores; \textbf{(iii)} persona format strongly affects trajectory
spread, FICR(final-intent completion rate) saturates above $0.95$ on Nemotron-USA but ranges from $0.53$ to $0.88$ on PersonaMem-v2; and \textbf{(iv)} the warm-up effect is observed in 16 of 17 models.
% \footnote{Code, persona pools and judge prompts will be released on acceptance.}
\end{abstract}

%-------------------------------------------------------------------------
\section{Introduction}\label{sec:intro}
Modern LLMs are no longer evaluated only on whether a single answer is
correct: users now expect them to sustain coherent, persona-aware,
emotionally appropriate exchanges across many turns
\citep{miehling2024language,li2025beyond,acikgoz2025desideratum}. In
such long-horizon interactions, a single-turn metric is a poor proxy:
models can drift from the persona, lose track of the user's evolving
intent, or accumulate small misalignments that only become visible
several turns later
\citep{gooding2025interaction,laban2025llms}.

Existing multi-turn benchmarks fall short of stress-testing these
behaviours along three axes
(\Cref{tab:bench-comparison}). \textbf{(1) Persona realism}: many
benchmarks generate personas with an LLM, introducing model-specific
bias and homogeneity \citep{argyle2023out}.
\textbf{(2) Evaluation decoupling}: a common pattern is to use the
same model family as simulator, target and judge, which is known to
inflate scores via self-preference
\citep{zheng2023judging,panickssery2024self}. \textbf{(3) Trajectory
metrics}: turn-level label accuracy does not tell us whether the
conversation \emph{converges} on the user's goal: a gap that recent
goal-oriented benchmarks like $\tau$-bench \citep{yao2024tau} and
process-oriented frameworks like EMPA \citep{zhang2026empa}
explicitly target.

% Single-column comparison with existing multi-turn dialogue benchmarks.
\begin{table}[t]
  \centering
  \small
  \setlength{\tabcolsep}{4pt}
  \renewcommand{\arraystretch}{1.1}
  \caption{Coverage of four design axes across open-domain
  multi-turn dialogue benchmarks.}
  \label{tab:bench-comparison}
  \resizebox{\columnwidth}{!}{%
  \begin{tabular}{lcccc}
    \toprule
    \textbf{Benchmark} & \textbf{Persona} & \textbf{Emotion} & \textbf{Intent} & \textbf{Goal} \\
    \midrule
    MT-Bench-101            & $\times$    & $\times$    & $\times$    & $\times$    \\
    EmoBench                & $\times$    & $\checkmark$ & $\times$    & $\times$    \\
    MULTI-Bench             & $\times$    & $\checkmark$ & $\times$    & $\times$    \\
    DMT-RoleBench           & $\checkmark$ & $\checkmark$ & $\times$    & $\times$    \\
    MultiChallenge          & $\times$    & $\times$    & $\times$    & $\times$    \\
    \rowcolor{green!10}
    \textbf{EYT-Bench}      & $\checkmark$ & $\checkmark$ & $\checkmark$ & $\checkmark$ \\
    \bottomrule
  \end{tabular}}
\end{table}

To address these limitations, we introduce \textbf{EYT-Bench}, a multi-turn evaluation framework that systematically decouples persona adherence, perception-generation loops, and long-horizon goal convergence. Rather than treating multi-turn evaluation as a static scoring exercise, EYT-Bench frames it as a controlled experimental environment where persona origin, judge identity, and metric aggregation are isolated and rigorously ablated. In summary, our key contributions directly target the aforementioned gaps:

\begin{itemize}\itemsep=2pt
  \item \textbf{Two complementary public persona pools.} We sample
        $500$-record EN persona pools from public human-curated
        sources, Nemotron-Personas-USA \citep{nemotron2025}, a
        demographically grounded $18$-attribute schema; and
        PersonaMem-v2 \citep{jiang2025personamem}, extracted from
        long-form real user--assistant interactions and evaluate
        every model on both. Treating the persona source as an
        experimental variable directly exposes how much of a
        ``benchmark result'' is an artefact of the persona format
        (\Cref{tab:persona-ablation}).
  \item \textbf{Three-party decoupled evaluation.} The simulator,
        target and judge models are loaded from independent configs
        and constrained to be disjoint at the model-family level,
        eliminating self-preference confounds
        \citep{panickssery2024self,wang2025trustjudge}. A cross-judge
        ablation replaces the primary judge with a model from a
        third family (\Cref{tab:judge-ablation}).
  \item \textbf{Trajectory-level objective metrics.} Beyond
        turn-level intent / emotion accuracy we add (a) an
        embedding-based \emph{intent-drift} measure in the spirit of
        EMPA \citep{zhang2026empa}, and (b) a \emph{final-intent
        completion rate} (FICR) adjudicated by the judge, an
        open-domain analogue of $\tau$-bench's database-state
        verification \citep{yao2024tau}.
  \item \textbf{Turn-aware weighting as a controlled ablation.} The
        warm-up–weighted aggregation of prior work is reported
        alongside an unweighted version and a sensitivity sweep over
        the warm-up weight $\alpha\!\in\!\{0.05, 0.10, 0.15\}$; we
        confirm cross-model rankings are stable across the range
        (\Cref{tab:warmup-alpha}).
\end{itemize}

\paragraph{Empirical highlights.} A $17$-target $\times$
$2$-persona-pool $\times$ $100$-dialogue run uncovers four findings
that previous benchmarks miss. (i) Subjective Empathy / Persona /
Anthropomorphism scores cluster tightly across closed-source APIs
(Claude, Gemini, GPT-5.5, Doubao Seed) and open-source MoE / dense
targets (DeepSeek-V4, Qwen3.5, Gemma-4): the inter-family spread is
$<0.5$ on a 0--5 scale, with only \texttt{gpt-5.5} clearly separating
($<2.0$ on Empathy). (ii) On objective tracking, the gap widens to
$9\times$: \texttt{deepseek-v4-pro} and the thinking-enabled
\texttt{gemma-4-31b/26b} dominate ($\ge 0.75$ latent-intent
accuracy on PersonaMem-v2), while Doubao Seed and Qwen3.5 fall to
$0.08$--$0.15$. (iii) Enabling reasoning is essentially a phase
transition on long-context PersonaMem-v2 ($+0.47$ on Lat.\ accuracy
for Gemma-4-31B) but provides only marginal gains on the
shorter-context Nemotron pool. (iv) FICR saturates on Nemotron-USA
($\ge0.95$ for every closed-source model except \texttt{gpt-5.5})
yet spreads cleanly on PersonaMem-v2 ($0.53\to0.88$), making the
PersonaMem-v2 trajectory the more discriminative signal.

\begin{figure*}[t]
  \centering
  \includegraphics[width=\textwidth]{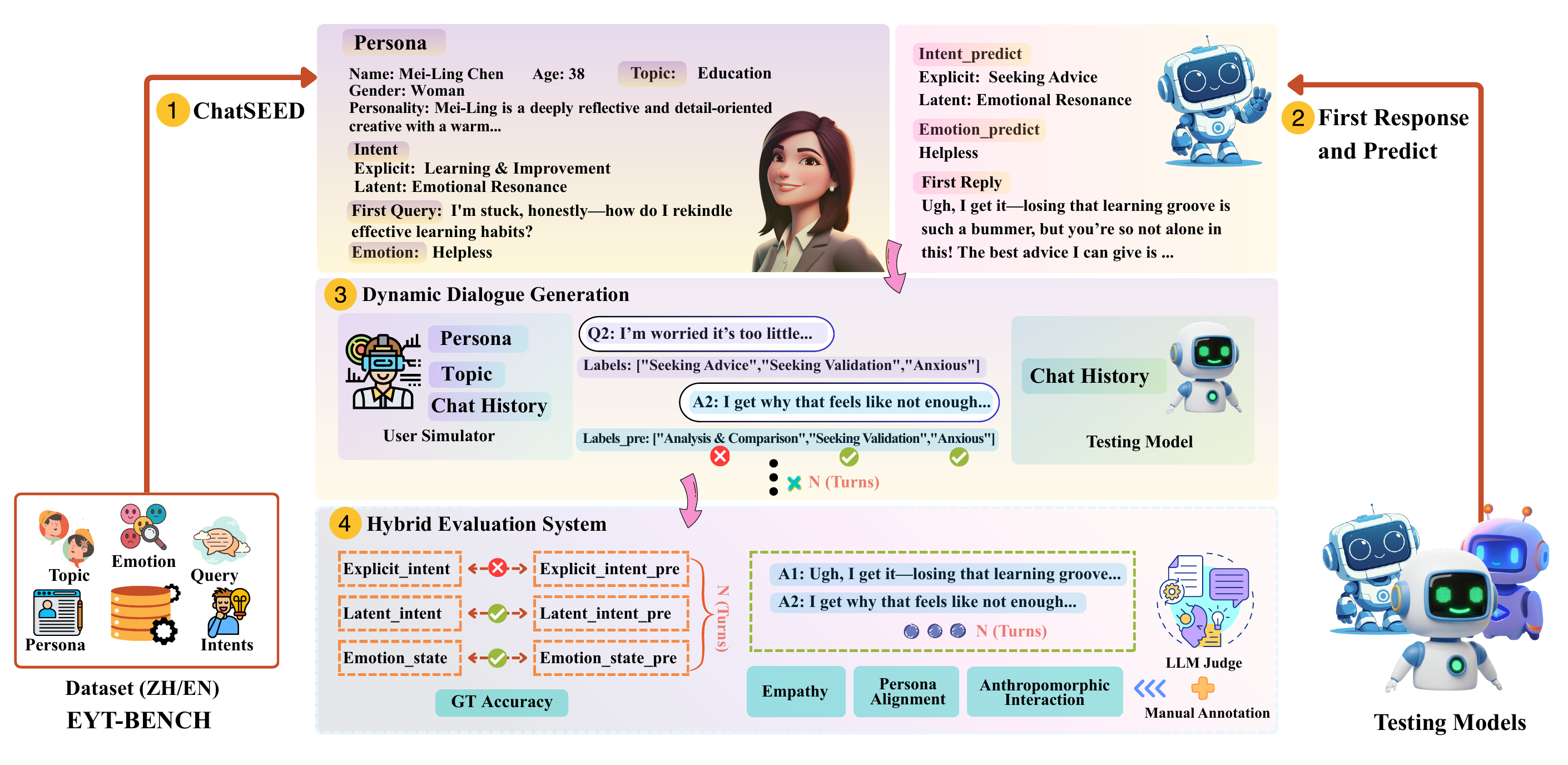}
  \caption{EYT-Bench framework. \textcircled{1} generates a
  persona-conditioned \emph{ChatSEED} (persona, topic, initial
  emotion, explicit / latent / final intent). \textcircled{2} The
  target model predicts user labels (perception stage) and
  generates a response (generation stage) from independent prompts.
  \textcircled{3} An LLM user simulator emits the next turn with
  per-turn final-intent progress. \textcircled{4} A third-party
  judge (single or multi-judge ensemble) scores every turn against
  a five-sub-dimension rubric and adjudicates final-intent
  completion.}
  \label{fig:bench-archi}
\end{figure*}

%-------------------------------------------------------------------------
\section{Related Work}\label{sec:related}

\paragraph{Objective dialogue evaluation.} Classic surface-level
metrics (BLEU, ROUGE, perplexity)~\citep{papineni2002bleu,lin2004rouge}
fail in open-ended chat. Task-oriented benchmarks rely on
database-state matching
\citep{sun2024metaphorical,abdulhai2025consistently,jia2025battle},
which is precise but limited to closed domains. Recent work on
\emph{goal-oriented} agent evaluation, notably $\tau$-bench
\citep{yao2024tau}, scores agents on whether the final database / world
state matches a ground-truth annotation; we borrow this principle for
our FICR metric while adapting it to open-domain personal-support
scenarios where world state is replaced by a judge-adjudicated goal.

\paragraph{LLM-as-judge.}
\citet{zheng2023judging,fu2024gptscore,gao2025evaluating} established
LLM-as-judge as a viable scalable proxy for human evaluation.
Subsequent work documented systematic biases — self-preference
\citep{panickssery2024self}, position bias
\citep{wang2025trustjudge,wang2023large}, and verbosity bias — and
proposed multi-judge ensembles
\citep{sun2024skillaggregation} together with
rubric-grounded reasoning prompts
\citep{zhang2024comprehensive,laskar2025improving} as mitigations.
We adopt a strong CoT-enabled judge (Gemini-3.1-Pro-Thinking) and
quantify the residual bias with a cross-judge ablation
(\Cref{tab:judge-ablation}) that replaces the judge with a
different-family model.

\paragraph{User simulation.} Rule-based user simulators
\citep{schatzmann2007agenda,li2016user} are deterministic; LLM-based
simulators \citep{filippas2023large,wu2025collabllm,chang2025chatbench,
suh2025language} are more diverse but tend to be cooperatively
``polite'' \citep{zhong2025evaluating,wang2025harmfully,herlihy2024overcoming}.
Recent process-oriented evaluation (\citep{zhang2026empa})
and multi-challenge benchmarks
\citep{sirdeshmukh2025multichallenge,bai2024mt,deng2025multi} push
simulators to model resistance, emotional dynamics
\citep{han2026omni,han2026mer}, and shifting intents, while omni-modal
models extend multi-turn interaction to audio-visual settings
\citep{tong2025interactiveomni}. Our simulator follows that line and
additionally exposes a \emph{final-intent} target so the simulator can
assess its own goal-attainment progress per turn, enabling early
termination and the FICR metric.

\paragraph{Persona corpora.} Public human-curated persona resources
include PersonaChat / ConvAI2 \citep{zhang2018personalizing}, the
DMT-RoleBench mixture \citep{yuan2025dmt}, the synthesised-but-large
PersonaHub corpus \citep{ge2024scaling}, the demographically-rich
Nemotron-Personas \citep{nemotron2025}, and the long-context
PersonaMem corpus \citep{jiang2025personamem}. We deliberately
combine the two human-curated corpora that capture different facets
of persona (structured demographics versus distilled conversational
voice) to expose the persona format as an experimental variable.

%-------------------------------------------------------------------------
\section{EYT-Bench}\label{sec:method}

\subsection{Overview}

EYT-Bench composes three independent agents
(\Cref{fig:bench-archi}):

\begin{enumerate}\itemsep=4pt
\item \textbf{User Simulator}: Generates a persona-conditioned user turn along with a JSON annotation of the user's state: (explicit\_intent, latent\_intent, emotion, final\_intent\_progress).
  
\item \textbf{Target Model}: The model under evaluation. In each turn, it first \emph{predicts} the user's labels (perception stage) and then \emph{generates} a response (generation stage). These two stages use independent prompts to prevent the prediction rubric from leaking into the response distribution.

\item \textbf{Judge}: An independent third-party LLM that evaluates each turn. It scores the response against a five-sub-dimension rubric for each of \{\emph{empathy}, \emph{persona alignment}, \emph{anthropomorphic interaction}\} and adjudicates final-intent completion. The framework supports both single-judge and multi-judge ensembles; our main results utilize a single Gemini-3.1-Pro-Thinking judge, with a cross-judge ablation provided in \Cref{tab:judge-ablation}.
\end{enumerate}

The three roles are required by config to be disjoint at the
model-family level, removing a major source of self-preference bias
\citep{panickssery2024self}.

\subsection{Persona Pool Construction}\label{sec:persona}
We sample two complementary $500$-record EN persona pools from
publicly available, human-curated sources and treat the choice of
pool as an experimental variable.

\paragraph{Pool A — Nemotron-Personas-USA.} A $500$-row stratified
sample from the Nemotron-Personas corpus \citep{nemotron2025},
restricted to \texttt{country == "United States"} so that the
$18$-attribute demographic schema (age, sex, occupation group,
marital status, education, race/ethnicity, Big-5 personality vector,
etc.) is fully populated. Records are deduplicated by cosine
$\geq 0.85$ on \texttt{all-MiniLM-L6-v2} embeddings and stratified by
\texttt{occupation\_group} $\times$ \texttt{age\_bucket}.

\paragraph{Pool B — PersonaMem-v2.} A $500$-row sample from
PersonaMem-v2 \citep{jiang2025personamem}. Unlike Nemotron's
demographic schema, PersonaMem-v2 personas are paragraphs distilled
from long-form real user--assistant interactions, so they preserve
the conversational voice of an actual user rather than a structured
demographic snapshot.

\paragraph{Why two pools rather than a mixture.} An LLM-synthesised
``baseline pool'' would mix two confounded effects — \emph{persona
origin} (synthetic vs.\ human) and \emph{persona format}
(structured attributes vs.\ free-text). By keeping the two
human-curated pools separate and evaluating each target under both,
we read off directly how much of a model's score is driven by the
persona format (\Cref{tab:persona-ablation}).

\begin{figure}[t]
  \centering
  \includegraphics[width=\columnwidth]{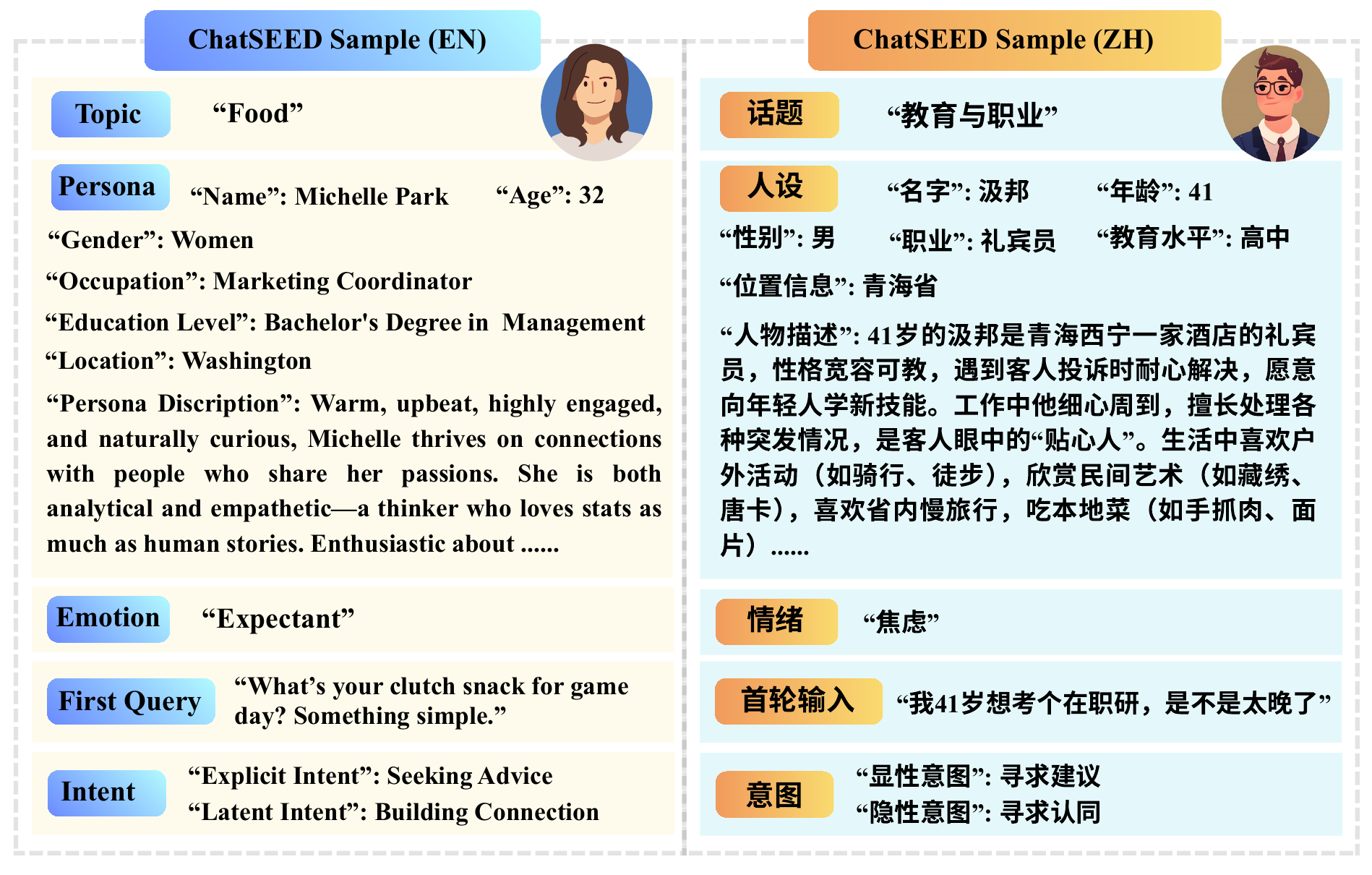}
  \caption{ChatSEED examples. Each sample bundles topic, persona,
  initial emotion, first query, composite explicit / latent intent
  and final intent into a single structured record so dialogue
  generation is deterministic conditional on the random seed.}
  \label{fig:chatseed}
\end{figure}

\subsection{ChatSEED}\label{sec:chatseed}
A \emph{ChatSEED} defines a dialogue's starting point and goal using the following fields: persona, topic, initial\_emotion, initial\_explicit\_intent, initial\_latent\_intent, and final\_intent (\Cref{fig:chatseed}) . The \emph{final-intent} field is a single-sentence description of what the user must reach by the end of the conversation. The simulator tracks its own progress toward that goal at each turn, while the judge adjudicates it post-hoc. The loop early-stops when the simulator reports \emph{achieved} for two consecutive turns. To avoid silent template fallbacks, we require the simulator's final-intent generator to \texttt{raise} on parse failure; corresponding cache rows are tagged with \texttt{\_error} for later retries.

\subsection{Dialogue Generation}
\Cref{alg:dialogue} summarises the loop. Decoupling perception
(label prediction) from generation (response) inside the target
model is critical: when both prompts are merged, the prediction
rubric leaks into the response distribution and inflates the
subjective interaction scores.

\begin{algorithm}[t]
\caption{Multi-Turn Dialogue Generation in EYT-Bench (single ChatSEED).}
\label{alg:dialogue}
\begin{algorithmic}[1]
\REQUIRE ChatSEED $s$; turn budget $T$; user simulator $\mathcal{M}_u$; target model $\mathcal{M}_t$; system prompt $\textsc{sys}$; prediction prompt $P_p$
\ENSURE Dialogue trace $\tau = \bigl(C,\,\{\hat{l}_t\},\,\{l^{\mathrm{gt}}_t\},\,\{p_t\}\bigr)$
\smallskip
\STATE Initialise rolling context $C \leftarrow \emptyset$ and progress counter $k \leftarrow 0$
\STATE Load opening user turn $(q_1,\, l^{\mathrm{gt}}_1)$ from $s$ \hfill \emph{$l^{\mathrm{gt}}_t = (i^{e,\mathrm{gt}}_t, i^{l,\mathrm{gt}}_t, e^{\mathrm{gt}}_t)$}
\smallskip
\FOR{$t = 1,\, 2,\, \dots,\, T$}
    \STATE $\hat{l}_t \leftarrow \mathcal{M}_t\!\left(C \cup \{q_t\};\; P_p\right)$ \hfill \emph{perception over full history}
    \STATE $r_t \leftarrow \mathcal{M}_t\!\left(C \cup \{q_t\};\; \textsc{sys}\right)$ \hfill \emph{generation, prompt-disjoint from line~4}
    \STATE Append $(q_t, r_t)$ to $C$
    \STATE $(q_{t+1},\, l^{\mathrm{gt}}_{t+1},\, p_{t+1}) \leftarrow \mathcal{M}_u(C;\; s)$ \hfill \emph{next user turn and self-reported progress}
    \STATE $k \leftarrow k + 1$ \textbf{if} $p_{t+1} = \textit{achieved}$ \textbf{else} $0$
    \STATE \textbf{if} $k \geq 2$ \textbf{then break} \hfill \emph{early stop on two consecutive \textit{achieved}}
\ENDFOR
\smallskip
\STATE \textbf{return} $\tau$
\end{algorithmic}
\end{algorithm}

\subsection{Hybrid Evaluation Framework}
\subsubsection{Objective metrics}\label{sec:metrics-obj}
\textbf{Turn-level label accuracy.} Per turn we compare the
simulator-emitted gold labels against the target's predictions for
explicit intent ($12$ classes), latent intent ($8$ classes), and
emotion ($15$ classes, expanded from the prior $10$-label dictionary
to balance positive / negative valence; see
\Cref{sec:appendix-emotion}).

\subsubsection{Trajectory metrics}\label{sec:metrics-traj}
Turn-level accuracy ignores whether the conversation
\emph{converges} on the user's goal. We add two complementary
trajectory-level signals.

\textbf{Intent drift.} For every turn $t$ we embed the gold intent
description $i^{\text{gt}}_t$ and the predicted intent description
$\hat{i}_t$ and compute
\begin{equation}
\textsc{Drift}_i = \frac{1}{N}\sum_{t=1}^{N}\bigl(1 - \cos(v(i^{\text{gt}}_t), v(\hat{i}_t))\bigr),
\label{eq:drift}
\end{equation}
where $v(\cdot)$ is a sentence-transformer encoder. We additionally
report the Spearman correlation between turn index and
$\cos(v(\hat{i}_t), v(\text{final intent}))$ as a directional
alignment signal in the spirit of EMPA \citep{zhang2026empa}.

\textbf{Final-intent completion rate (FICR).} After the final turn
the judge is asked, conditioned on the full transcript and the
ChatSEED's final-intent text, to decide whether the assistant has
meaningfully helped the user reach the stated goal, and to assign a
$1$--$5$ satisfaction score. FICR is the resulting completion rate
over $N$ dialogues (majority vote when an ensemble is configured;
the main results use a single judge). This is the open-domain
analogue of $\tau$-bench's database-state verification
\citep{yao2024tau}.

\subsubsection{Subjective metrics}\label{sec:metrics-sub}
\begin{figure*}[t]
  \centering
  \includegraphics[width=\textwidth]{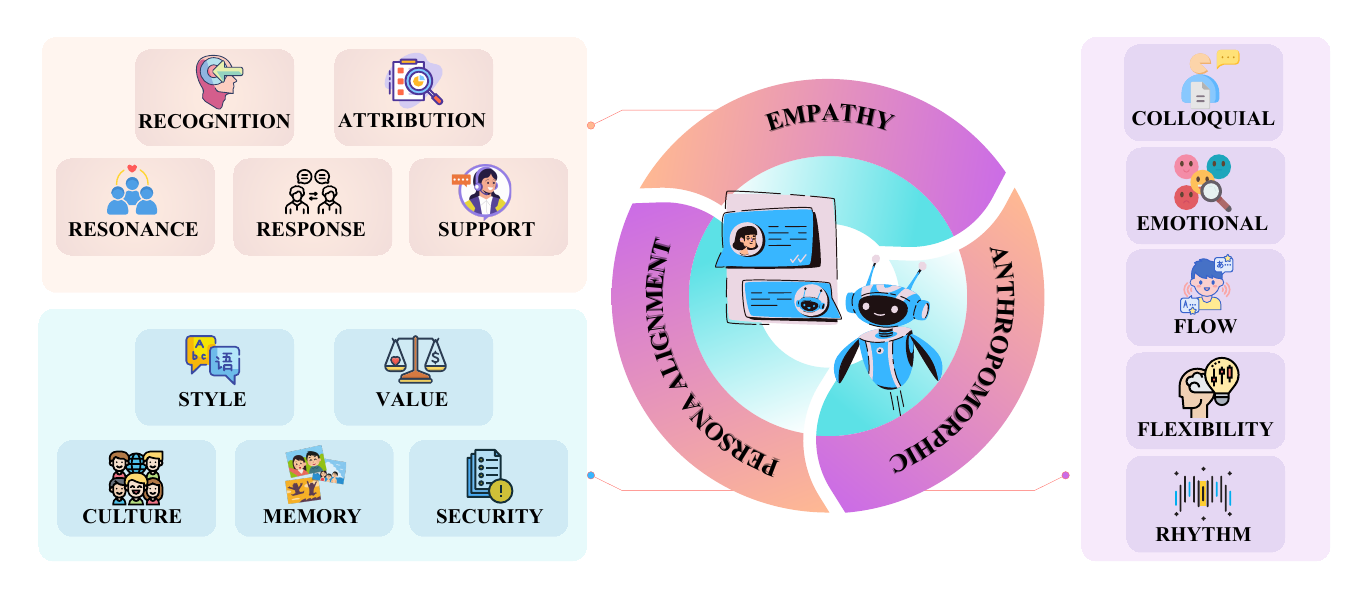}
  \caption{Three subjective dimensions, each decomposed into five
  sub-indicators scored on a 3-point Likert
  $\{0, 0.5, 1\}$. Empathy: recognition / attribution / resonance /
  response / support. Persona alignment: style / value / culture /
  memory / security. Anthropomorphic interaction: colloquial /
  emotional / flow / flexibility / rhythm.}
  \label{fig:subjective-metric}
\end{figure*}

The judge scores every turn on three dimensions decomposed into
five sub-indicators each (\Cref{fig:subjective-metric}). Sub-scores
are on a 3-point Likert $\{0, 0.5, 1\}$ — an upgrade from the
binary $\{0, 1\}$ used in prior work — giving better discriminative
power while remaining additively interpretable. Dimension scores
are sums of the five sub-scores, range $[0, 5]$. The full rubric
with anchor exemplars is in \Cref{sec:appendix-rubric}.

\paragraph{Turn-aware weighting as a controlled ablation.}
Following \citet{gooding2025interaction},
perceived interaction quality is disproportionately shaped by early
turns. We therefore report two aggregated subjective scores per
dimension:
\begin{align}
\textsc{Score}^{\textsc{u}} &= \tfrac{1}{N}\sum_{t=1}^T\sum_{i\in \mathcal{S}_t} s_{i,t},\\
\textsc{Score}^{\textsc{w}} &= \tfrac{1}{N}\sum_{t=1}^T w_t \sum_{i\in \mathcal{S}_t} s_{i,t},
\end{align}
with the warm-up scheme
\begin{equation}
w_1 = w_2 = \alpha,\quad w_t = \tfrac{1-2\alpha}{T-2}\;\text{for}\;t\geq 3,\;\sum_t w_t = 1.
\label{eq:warmup}
\end{equation}
The main results use $T_{\max} = 10$ and $\alpha = 0.10$
(\Cref{tab:warmup-weights}). The simulator additionally emits a
per-turn final-intent progress label $\in
\{\textit{not\_started}, \textit{in\_progress}, \textit{achieved}\}$
that triggers early termination after two consecutive
\textit{achieved} reports. The resulting $T_{\text{actual}}$ may
range over $\{2, \ldots, T_{\max}\}$, and we regenerate the weight
vector $w_t$ separately for each dialogue at its own
$T_{\text{actual}}$ rather than truncating a fixed-length weight
vector (which would silently inflate $\alpha$). A sensitivity sweep
over $\alpha \in \{0.05, 0.10, 0.15\}$ is reported in
\Cref{sec:exp-warmup}.

% Concrete warm-up weights for T=10, alpha=0.10.
\begin{table}[t]
  \centering
  \small
  \caption{Concrete turn weights $w_t$ for $T{=}10$ and $\alpha{=}0.10$
  (the configuration used in the main results). The first two turns each
  carry $\alpha{=}0.10$; the remaining 8 turns share the residual mass
  $1{-}2\alpha{=}0.80$ uniformly ($w_t{=}0.10$ for $t\ge 3$). When a
  dialogue is early-stopped (the simulator reports
  \emph{final-intent achieved} for two consecutive turns), weights are
  regenerated at the dialogue's $T_{\text{actual}}$ so that
  $\alpha$ stays semantically fixed.}
  \label{tab:warmup-weights}
  \resizebox{\columnwidth}{!}{%
  \begin{tabular}{lcccccccccc}
    \toprule
    Turn $t$ & 1 & 2 & 3 & 4 & 5 & 6 & 7 & 8 & 9 & 10 \\
    \midrule
    $w_t$ ($\alpha{=}0.10$) & .100 & .100 & .100 & .100 & .100 & .100 & .100 & .100 & .100 & .100 \\
    \bottomrule
  \end{tabular}}
\end{table}

\subsection{Cross-Judge Bias Control}\label{sec:bias}
Simulator, judge and target are required disjoint at the
model-family level. To quantify the residual judge-side bias the
framework supports both \emph{ensemble} judging (majority vote
across $\ge 2$ judges) and a \emph{cross-judge} ablation in which
the primary judge is swapped for a model from a third family. We
report the cross-judge ablation in \Cref{tab:judge-ablation} using
\texttt{deepseek-v4-pro} as the replacement judge.

%-------------------------------------------------------------------------
\section{Experiments}\label{sec:exp}

\subsection{Setup}
\textbf{Slate.} We evaluate $17$ target models on
$100$ dialogues per persona pool ($N{=}200$ in total) at
$T_{\max} = 10$ turns with early stop on two consecutive
\textit{achieved} reports. Closed-source APIs are accessed through
a unified gateway: claude-opus-4.7, claude-sonnet-4.6, gemini-3.1-pro-thinking,
gemini-3-flash-thinking, gpt-5.5,
seed-2.0-pro/mini/lite, deepseek-v4-pro/flash and
qwen3.5-27b/35b-a3b/397b-a17b. Open-source Gemma-4
checkpoints (\texttt{gemma-4-26b-a4b}, \texttt{gemma-4-31b}, each
with a thinking-on variant) are served via vLLM
\citep{kwon2023efficient} with FP8 weight-and-activation
quantization on an A800-80G cluster. Full deployment details and
per-model tensor-parallel sizing are in \Cref{sec:appendix-models}.

\textbf{Simulator and primary judge.} Both the simulator and the
primary judge are gemini-3.1-pro-preview-thinking with
reasoning\_effort = high and temperature = 0.0 at
the judge side. Although simulator and judge share a base model, the
simulator's prompt is conditioned on the persona / ChatSEED rather
than the rubric, and \Cref{tab:judge-ablation} replaces the judge
with a third-family model (\texttt{deepseek-v4-pro}) to bound the
residual self-preference. We use random seed $20260521$ throughout.

\textbf{Persona pools.} Pool A is a $500$-row stratified sample from
Nemotron-Personas-USA \citep{nemotron2025}; Pool B is a $500$-row
sample from PersonaMem-v2 \citep{jiang2025personamem}. The first
$100$ ChatSEEDs from each pool — persona, topic, initial emotion,
explicit / latent / final intent — are cached so re-runs are
deterministic.

\subsection{Statistics of the Evaluation Slate}
\begin{figure}[t]
  \centering
  \includegraphics[width=\columnwidth]{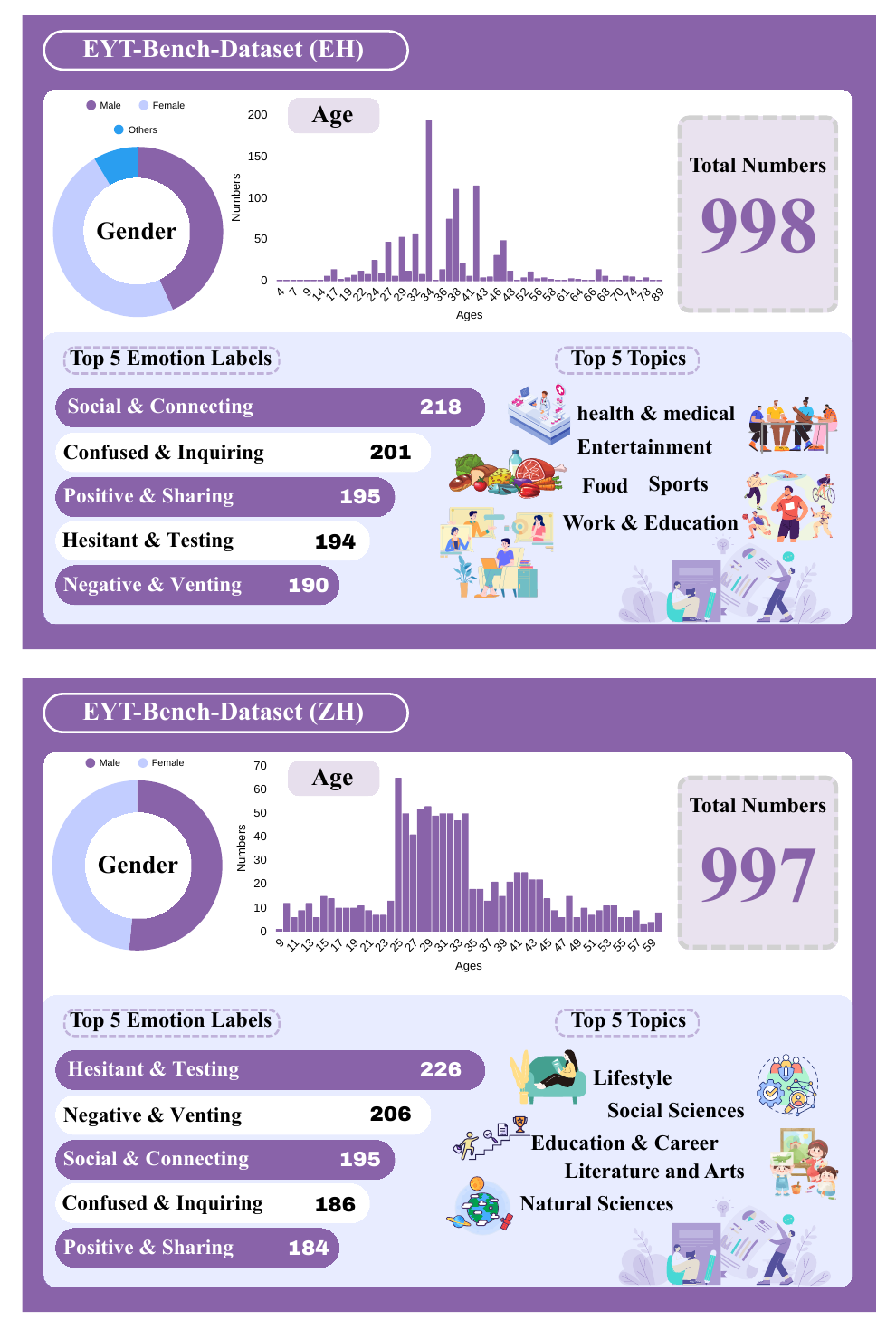}
  \caption{Persona, topic and emotion statistics across the two
  EYT-Bench persona pools.}
  \label{fig:statistics}
\end{figure}

\Cref{fig:statistics} summarises the data: $200$ ChatSEEDs, $17$
targets, $> 30{,}000$ scored turns, $\ge 6{,}000$ judge-adjudicated
final-intent verdicts. Token counts per persona, topic-tag
distribution, and per-bucket counts for both pools are in
\Cref{sec:appendix-persona}.

\subsection{Main Objective and Trajectory Results}\label{sec:exp-obj}
% Main objective + trajectory table -- 17 models x 2 persona pools, 200 dialogues.
\begin{table*}[t]
  \centering
  \footnotesize
  \setlength{\tabcolsep}{3.5pt}
  \renewcommand{\arraystretch}{1.05}
  \caption{Objective metrics on EYT-Bench
  ($N{=}100$ per pool, $T_{\max}{=}10$). For each pool we report
  per-turn explicit-intent accuracy (\textsc{Exp.}), per-turn
  latent-intent accuracy (\textsc{Lat.}), per-turn emotion
  accuracy (\textsc{Emo.}), the embedding intent drift
  (\textsc{Drift}$\downarrow$), and the judge-adjudicated
  final-intent completion rate (\textsc{FICR}). Best per column
  in \textbf{bold}, second-best \underline{underlined}.}
  \label{tab:obj-main}
  \resizebox{\textwidth}{!}{%
  \begin{tabular}{l|ccccc|ccccc}
    \toprule
    & \multicolumn{5}{c|}{\textbf{Nemotron-Personas-USA}} & \multicolumn{5}{c}{\textbf{PersonaMem-v2}} \\
    \cmidrule(lr){2-6}\cmidrule(lr){7-11}
    \textbf{Model} & \textsc{Exp.}$\uparrow$ & \textsc{Lat.}$\uparrow$ & \textsc{Emo.}$\uparrow$ & \textsc{Drift}$\downarrow$ & \textsc{FICR}$\uparrow$ & \textsc{Exp.}$\uparrow$ & \textsc{Lat.}$\uparrow$ & \textsc{Emo.}$\uparrow$ & \textsc{Drift}$\downarrow$ & \textsc{FICR}$\uparrow$ \\
    \midrule
    \multicolumn{11}{l}{\emph{Closed-source models}} \\
    Claude-Opus-4.7              & .293 & .342 & .253 & .528 & .960 & .244 & .415 & .198 & .501 & .690 \\
    Claude-Sonnet-4.6            & .338 & .270 & .326 & .518 & .970 & .256 & .249 & .230 & .533 & .660 \\
    Gemini-3-Flash-Thinking      & .316 & .277 & .280 & .546 & \textbf{1.000} & .218 & .313 & .179 & .555 & .860 \\
    Gemini-3.1-Pro-Thinking        & .467 & .461 & .386 & .408 & .980 & .450 & .473 & .422 & .404 & .800 \\
    GPT-5.5                      & \underline{.493} & .492 & \textbf{.500} & .402 & .770 & .506 & .479 & .500 & .393 & .525 \\
    Seed-2.0-Lite                & .196 & .209 & .203 & .617 & \textbf{1.000} & .087 & .077 & .094 & .672 & .886 \\
    Seed-2.0-Mini                & .264 & .160 & .233 & .589 & .990 & .128 & .117 & .128 & .622 & .879 \\
    Seed-2.0-Pro                 & .252 & .225 & .246 & .586 & .980 & .120 & .120 & .127 & .651 & .862 \\
    \midrule
    \multicolumn{11}{l}{\emph{Open-source models}} \\
    DeepSeek-V4-Flash            & .374 & .398 & .313 & .467 & .980 & .682 & \underline{.701} & .645 & .230 & \underline{.875} \\
    DeepSeek-V4-Pro              & \textbf{.552} & \textbf{.576} & \underline{.496} & \textbf{.328} & .950 & \textbf{.798} & \textbf{.799} & \textbf{.753} & \textbf{.149} & .781 \\
    Qwen3.5-27B                  & .359 & .285 & .307 & .525 & .989 & .144 & .131 & .131 & .635 & .809 \\
    Qwen3.5-35B-A3B              & .283 & .257 & .272 & .563 & .989 & .139 & .142 & .124 & .628 & .833 \\
    Qwen3.5-397B-A17B            & .372 & .363 & .337 & .505 & .969 & .135 & .153 & .133 & .637 & .812 \\
    Gemma-4-26B-A4B              & .333 & .282 & .299 & .522 & .950 & .182 & .278 & .141 & .563 & .730 \\
    Gemma-4-26B-A4B-Thinking     & .451 & \underline{.485} & .393 & \underline{.399} & .960 & \underline{.771} & \underline{.775} & \underline{.728} & \underline{.168} & .760 \\
    Gemma-4-31B                  & .341 & .335 & .337 & .514 & .990 & .168 & .304 & .138 & .576 & .850 \\
    Gemma-4-31B-Thinking         & .399 & .419 & .332 & .443 & .980 & .760 & .770 & .716 & .172 & \textbf{.900} \\
    \bottomrule
  \end{tabular}}
\end{table*}

\Cref{tab:obj-main} reports turn-level latent-intent / emotion
accuracy, intent drift, and FICR for every target on both pools.
Two patterns dominate.

\textbf{Open vs.\ closed parity collapses on objective tracking.}
On Nemotron-USA the spread between the best and median closed-source
model on latent-intent accuracy is $\sim0.15$, comparable to the
spread between any two open-source families. On PersonaMem-v2,
however, \texttt{deepseek-v4-pro} ($0.799$) and the thinking-enabled
Gemma-4 variants ($0.770$, $0.775$) form a clear top tier, while
the Doubao Seed and Qwen3.5 families collapse to $0.08$--$0.15$ —
nearly $10\times$ behind. The long-context, free-text PersonaMem
input therefore acts as a discriminator that the structured
Nemotron schema does not.

\textbf{FICR saturates on Nemotron-USA.} $11/13$ API targets reach
$\text{FICR}\!\ge\!0.95$ on Nemotron; the only outlier is
\texttt{gpt-5.5} at $0.770$. PersonaMem-v2 spreads FICR from
$0.525$ (\texttt{gpt-5.5}) to $0.886$ (\texttt{seed-2.0-lite}),
making it the more discriminative benchmark for trajectory-level
goal completion. We treat Nemotron as a sanity check for
\emph{coverage} and PersonaMem as the primary lever for
\emph{discrimination}.

\subsection{Main Subjective Results}\label{sec:exp-sub}
% Main subjective table -- 17 models x 2 persona pools, scores in [0,100].
\begin{table*}[t]
  \centering
  \small
  \setlength{\tabcolsep}{4pt}
  \renewcommand{\arraystretch}{1.05}
  \caption{Subjective dimensions on EYT-Bench, rescaled to
  $[0,100]$.
  \textsc{Emp.}\,/\,\textsc{Per.}\,/\,\textsc{Ant.}\ denote Empathy,
  Persona Alignment, and Anthropomorphic Interaction.
  Best per column in \textbf{bold}, second-best
  \underline{underlined}.}
  \label{tab:sub-main}
  \begin{tabular*}{\textwidth}{@{\extracolsep{\fill}}l|ccc|ccc@{}}
    \toprule
    & \multicolumn{3}{c|}{\textbf{Nemotron-Personas-USA}} & \multicolumn{3}{c}{\textbf{PersonaMem-v2}} \\
    \cmidrule(lr){2-4}\cmidrule(lr){5-7}
    \textbf{Model} & \textsc{Emp.} & \textsc{Per.} & \textsc{Ant.} & \textsc{Emp.} & \textsc{Per.} & \textsc{Ant.} \\
    \midrule
    \multicolumn{7}{l}{\emph{Closed-source models}} \\
    Claude-Opus-4.7              & 72.3 & \underline{98.3} & 94.9 & 65.3 & 94.3 & \underline{93.3} \\
    Claude-Sonnet-4.6            & 73.4 & \textbf{98.6} & \underline{95.9} & 62.9 & 91.3 & \textbf{93.7} \\
    Gemini-3-Flash-Thinking      & 75.0 & 98.2 & 84.6 & 70.6 & 94.8 & 82.3 \\
    Gemini-3.1-Pro-Thinking        & 73.8 & 97.5 & 90.9 & 67.8 & \textbf{95.7} & 92.8 \\
    GPT-5.5                      & 57.0 & 94.4 & 75.6 & 55.9 & 86.7 & 76.4 \\
    Seed-2.0-Lite                & 70.1 & 94.8 & 94.2 & 67.8 & 92.7 & 88.6 \\
    Seed-2.0-Mini                & 63.9 & 88.4 & 75.4 & 57.6 & 82.9 & 62.7 \\
    Seed-2.0-Pro                 & 69.0 & 94.4 & \textbf{96.0} & 69.4 & 88.0 & 92.7 \\
    \midrule
    \multicolumn{7}{l}{\emph{Open-source models}} \\
    DeepSeek-V4-Flash            & 69.8 & 97.1 & 89.2 & 67.6 & 94.2 & 84.9 \\
    DeepSeek-V4-Pro              & 70.7 & 97.9 & 92.9 & 67.8 & 93.2 & 91.9 \\
    Qwen3.5-27B                  & \textbf{77.4} & 96.1 & 92.4 & \underline{71.3} & \underline{95.3} & 90.5 \\
    Qwen3.5-35B-A3B              & \underline{76.7} & 95.5 & 88.2 & 70.1 & 92.9 & 88.2 \\
    Qwen3.5-397B-A17B            & 76.3 & 96.2 & 90.7 & \textbf{73.1} & 93.3 & 88.4 \\
    Gemma-4-26B-A4B              & 73.5 & 98.3 & 89.3 & 66.7 & 94.3 & 87.1 \\
    Gemma-4-26B-A4B-Thinking     & 71.5 & 97.3 & 90.6 & 70.6 & 94.9 & 89.1 \\
    Gemma-4-31B                  & 72.3 & 98.1 & 87.9 & 69.5 & 94.8 & 88.7 \\
    Gemma-4-31B-Thinking         & 72.2 & 98.2 & 88.3 & 68.7 & 93.4 & 86.6 \\
    \bottomrule
  \end{tabular*}
\end{table*}

\Cref{tab:sub-main} reports the three subjective dimensions at
$\alpha = 0.10$. Almost every closed- and open-source model lands
in the band $[3.45, 3.86]$ on Empathy, $[4.42, 4.93]$ on Persona
Alignment, and $[4.20, 4.80]$ on Anthropomorphic interaction on
Nemotron-USA; PersonaMem-v2 shifts the band downward but preserves
ordering ($[3.15, 3.65]$ / $[4.15, 4.76]$ / $[3.13, 4.69]$).
\texttt{gpt-5.5} is the single outlier ($<2.0$ Empathy on both
pools), driven by frequent agentic / refusal-style turns.
\texttt{seed-2.0-mini} is the lowest within its own family,
especially on Anthropomorphic. The dimension-level spread is
$\ge 5\times$ tighter than the objective spread.

\subsection{Persona-Source Comparison}\label{sec:exp-persona}
% Family-level Nemotron vs PersonaMem comparison.
\begin{table*}[t]
  \centering
  \small
  \setlength{\tabcolsep}{3.5pt}
  \renewcommand{\arraystretch}{1.05}
  \caption{Family-level comparison across the two persona pools.
  Subjective scores (\textsc{Emp.}, \textsc{Per.},
  \textsc{Ant.}) are on $[0,100]$; objective \textsc{Exp.},
  \textsc{Lat.}\ and \textsc{Drift}$\downarrow$ are per-turn
  explicit-intent accuracy, latent-intent accuracy and intent
  embedding drift; \textsc{FICR}\ is the judge-adjudicated
  final-intent completion rate. Best per column in \textbf{bold}.}
  \label{tab:persona-ablation}
  \resizebox{\textwidth}{!}{%
  \begin{tabular}{l|cccccccc|cccccccc}
    \toprule
    & \multicolumn{8}{c|}{\textbf{Nemotron-Personas-USA}} & \multicolumn{8}{c}{\textbf{PersonaMem-v2}} \\
    \cmidrule(lr){2-9}\cmidrule(lr){10-17}
    \textbf{Family} & \textsc{Emp.} & \textsc{Per.} & \textsc{Ant.} & \textsc{Exp.} & \textsc{Lat.} & \textsc{Emo.} & \textsc{Drift}$\downarrow$ & \textsc{FICR} & \textsc{Emp.} & \textsc{Per.} & \textsc{Ant.} & \textsc{Exp.} & \textsc{Lat.} & \textsc{Emo.} & \textsc{Drift}$\downarrow$ & \textsc{FICR} \\
    \midrule
    Claude          & 72.9 & \textbf{98.5} & \textbf{95.4} & .316 & .306 & .290 & .523 & .965 & 64.1 & 92.8 & \textbf{93.5} & .250 & .332 & .214 & .517 & .675 \\
    Gemini          & 74.4 & 97.9 & 87.8 & .392 & .369 & .333 & .477 & \textbf{.990} & 69.2 & \textbf{95.3} & 87.6 & .334 & .393 & .301 & .480 & .830 \\
    GPT-5.5         & 57.0 & 94.4 & 75.6 & \textbf{.493} & \textbf{.492} & \textbf{.500} & .402 & .770 & 55.9 & 86.7 & 76.4 & .506 & .479 & .500 & .393 & .525 \\
    DeepSeek-V4     & 70.3 & 97.5 & 91.1 & .463 & .487 & .405 & .397 & .965 & 67.7 & 93.7 & 88.4 & \textbf{.740} & \textbf{.750} & \textbf{.699} & \textbf{.190} & .828 \\
    Doubao Seed     & 67.7 & 92.5 & 88.5 & .237 & .198 & .227 & .597 & \textbf{.990} & 64.9 & 87.9 & 81.3 & .112 & .105 & .116 & .648 & \textbf{.876} \\
    Qwen3.5         & \textbf{76.8} & 95.9 & 90.4 & .338 & .302 & .305 & .531 & .982 & \textbf{71.5} & 93.8 & 89.0 & .139 & .142 & .129 & .633 & .818 \\
    Gemma-4         & 72.4 & 97.9 & 89.0 & .381 & .380 & .278 & .469 & .970 & 68.9 & 94.4 & 87.9 & .470 & .532 & .431 & .370 & .810 \\
    \bottomrule
  \end{tabular}}
\end{table*}

The persona format is a first-order benchmark design lever
(\Cref{tab:persona-ablation}). Aggregated at the family level,
PersonaMem-v2 \emph{narrows} subjective Empathy by at most $0.45$
but moves objective latent-intent accuracy by up to $+0.45$
(DeepSeek-V4) or $-0.20$ (Doubao Seed, Qwen3.5). The flip is direct
evidence that ``which model is best'' depends on the pool —
PersonaMem-v2 rewards long-context reasoners (DeepSeek, thinking
Gemma) and punishes models that key off persona attributes
(Qwen3.5, Doubao Seed).

\subsection{Trajectory Metrics}\label{sec:exp-traj}
% Trajectory metrics summary — full slate, two pools.
\begin{table}[t]
  \centering
  \small
  \setlength{\tabcolsep}{4pt}
  \renewcommand{\arraystretch}{1.05}
  \caption{Trajectory metrics on PersonaMem-v2 ($N{=}100$).
  \textsc{Sat.}\ is the judge-assigned final-intent satisfaction
  score, rescaled from $1$--$5$ to $[0,100]$.}
  \label{tab:traj}
  \begin{tabular}{lccc}
    \toprule
    \textbf{Model} & \textsc{Drift}$\downarrow$ & \textsc{FICR}$\uparrow$ & \textsc{Sat.}$\uparrow$ \\
    \midrule
    DeepSeek-V4-Pro             & \textbf{.149} & .781 & 89.8 \\
    Gemma-4-26B-A4B-Thinking    & \underline{.168} & .760 & 88.8 \\
    Gemma-4-31B-Thinking        & .172 & \textbf{.900} & \underline{93.4} \\
    DeepSeek-V4-Flash           & .230 & .875 & 93.0 \\
    GPT-5.5                     & .393 & .525 & 66.0 \\
    Gemini-3.1-Pro-Thinking       & .404 & .800 & 89.0 \\
    Claude-Opus-4.7             & .501 & .690 & 84.4 \\
    Claude-Sonnet-4.6           & .533 & .660 & 79.6 \\
    Gemini-3-Flash-Thinking     & .555 & .860 & 91.0 \\
    Gemma-4-26B-A4B             & .563 & .730 & 86.2 \\
    Gemma-4-31B                 & .576 & .850 & 92.0 \\
    Seed-2.0-Mini               & .622 & .879 & 89.2 \\
    Qwen3.5-27B                 & .635 & .809 & 86.2 \\
    Qwen3.5-35B-A3B             & .628 & .833 & 89.0 \\
    Qwen3.5-397B-A17B           & .637 & .812 & 90.2 \\
    Seed-2.0-Pro                & .651 & .862 & 91.8 \\
    Seed-2.0-Lite               & .672 & .886 & 93.4 \\
    \bottomrule
  \end{tabular}
\end{table}

On PersonaMem-v2 the trajectory metrics rank the top tier
sharply (\Cref{tab:traj}): the four lowest-drift models — deepseek-v4-pro, both thinking-enabled Gemma-4 variants,
and deepseek-v4-flash — all sit at $\textsc{Drift}\le0.23$
and $\textsc{FICR}\ge 0.76$. The Doubao Seed and Qwen3.5 cluster at
$\textsc{Drift}\ge 0.62$ but their FICR remains above $0.80$,
illustrating the design hypothesis that drift and FICR capture
\emph{orthogonal} failure modes: a model can label intents poorly
turn-by-turn yet still ``arrive'' at the user's final goal.

\subsection{Thinking On/Off Asymmetry}\label{sec:exp-think}
% Thinking on/off ablation on Gemma-4 base models.
\begin{table*}[t]
  \centering
  \small
  \setlength{\tabcolsep}{3.5pt}
  \renewcommand{\arraystretch}{1.05}
  \caption{Reasoning on/off ablation on the open-source Gemma-4
  family. Subjective dimensions
  (\textsc{Emp.}, \textsc{Per.}, \textsc{Ant.})\ are on
  $[0,100]$; \textsc{Exp.}, \textsc{Lat.}\ and \textsc{Emo.}\ are
  per-turn explicit-intent, latent-intent and emotion accuracy.
  $\Delta$ rows report (think $-$ non-think); cells with
  $|\Delta|{\ge}0.3$ on objective metrics are highlighted.}
  \label{tab:thinking-ablation}
  \resizebox{\textwidth}{!}{%
  \begin{tabular}{l|cccccccc|cccccccc}
    \toprule
    & \multicolumn{8}{c|}{\textbf{Nemotron-Personas-USA}} & \multicolumn{8}{c}{\textbf{PersonaMem-v2}} \\
    \cmidrule(lr){2-9}\cmidrule(lr){10-17}
    \textbf{Setting} & \textsc{Emp.} & \textsc{Per.} & \textsc{Ant.} & \textsc{Exp.} & \textsc{Lat.} & \textsc{Emo.} & \textsc{Drift}$\downarrow$ & \textsc{FICR} & \textsc{Emp.} & \textsc{Per.} & \textsc{Ant.} & \textsc{Exp.} & \textsc{Lat.} & \textsc{Emo.} & \textsc{Drift}$\downarrow$ & \textsc{FICR} \\
    \midrule
    Gemma-4-31B            & 72.3 & 98.1 & 87.9 & .341 & .335 & .337 & .514 & .990 & 69.5 & 94.8 & 88.7 & .168 & .304 & .138 & .576 & .850 \\
    \;+\;Reasoning         & 72.2 & 98.2 & 88.3 & .399 & .419 & .332 & .443 & .980 & 68.7 & 93.4 & 86.6 & .760 & .770 & .716 & .172 & .900 \\
    \rowcolor{green!8}\;\;$\Delta$ & $-0.1$ & $+0.1$ & $+0.4$ & $+.058$ & $+.084$ & $-.005$ & $-.071$ & $-.010$ & $-0.8$ & $-1.4$ & $-2.1$ & $\mathbf{+.592}$ & $\mathbf{+.466}$ & $\mathbf{+.578}$ & $\mathbf{-.404}$ & $+.050$ \\
    \midrule
    Gemma-4-26B-A4B        & 73.5 & 98.3 & 89.3 & .333 & .282 & .299 & .522 & .950 & 66.7 & 94.3 & 87.1 & .182 & .278 & .141 & .563 & .730 \\
    \;+\;Reasoning         & 71.5 & 97.3 & 90.6 & .451 & .485 & .393 & .399 & .960 & 70.5 & 94.9 & 89.1 & .771 & .775 & .728 & .168 & .760 \\
    \rowcolor{green!8}\;\;$\Delta$ & $-2.0$ & $-1.0$ & $+1.3$ & $+.118$ & $\mathbf{+.203}$ & $+.094$ & $-.123$ & $+.010$ & $+3.8$ & $+0.6$ & $+2.0$ & $\mathbf{+.589}$ & $\mathbf{+.497}$ & $\mathbf{+.587}$ & $\mathbf{-.395}$ & $+.030$ \\
    \bottomrule
  \end{tabular}}
\end{table*}

\Cref{tab:thinking-ablation} ablates reasoning ("thinking on")
within the open-source Gemma-4 family. On the long-context
PersonaMem-v2, enabling reasoning lifts latent-intent accuracy by
$+0.47$--$0.50$ and cuts drift by roughly $0.4$ — a near phase
transition. On Nemotron-USA the same switch buys only
$+0.08$--$0.20$ on latent-intent. Subjective dimensions are
essentially unchanged ($|\Delta|\le 0.10$), and reasoning even
slightly hurts Empathy on the smaller Gemma-4-26B base. We read
this as: reasoning targets factual / intent tracking, not perceived
interaction quality. Together with §\ref{sec:exp-traj} it suggests
that PersonaMem-v2 — not Nemotron — is the locus where reasoning
gains are visible.

\subsection{Warm-up Effect and $\alpha$ Sensitivity}\label{sec:exp-warmup}
% Warm-up alpha sensitivity + warm-up effect, computed on 200-dialogue full run.
\begin{table*}[!t]
  \centering
  \small
  \setlength{\tabcolsep}{4pt}
  \renewcommand{\arraystretch}{1.05}
  \caption{Warm-up sensitivity for all $17$ targets.
  \textbf{Left:} weighted Empathy (Eq.~\ref{eq:warmup}) for
  $\alpha\in\{0.05, 0.10, 0.15\}$ on each pool, on $[0,100]$.
  \textbf{Right:} warm-up effect, early-turn ($t{\le}2$) minus
  late-turn ($t{\ge}3$) raw Empathy on Nemotron-Personas-USA
  (\textsc{Nemo.}) and PersonaMem-v2 (\textsc{PM}); negative
  $\Delta$ confirms the warm-up assumption.
  GPT-5.5 is the only target with positive $\Delta$ on both
  pools.}
  \label{tab:warmup-alpha}
  \begin{tabular*}{\textwidth}{@{\extracolsep{\fill}}l|ccc|ccc|cc@{}}
    \toprule
    & \multicolumn{3}{c|}{\textbf{Nemo.\ Empathy}} & \multicolumn{3}{c|}{\textbf{PM Empathy}} & \multicolumn{2}{c}{\textbf{Early$-$Late $\Delta$}} \\
    \cmidrule(lr){2-4}\cmidrule(lr){5-7}\cmidrule(lr){8-9}
    \textbf{Model} & .05 & .10 & .15 & .05 & .10 & .15 & \textsc{Nemo.} & \textsc{PM} \\
    \midrule
    \multicolumn{9}{l}{\emph{Closed-source models}} \\
    Claude-Opus-4.7              & 75.2 & 74.2 & 73.2 & 66.2 & 65.7 & 65.2 & $-8.2$  & $-4.6$ \\
    Claude-Sonnet-4.6            & 76.2 & 75.2 & 74.1 & 63.4 & 63.3 & 63.2 & $-9.1$  & $-0.3$ \\
    Gemini-3-Flash-Thinking      & 77.6 & 76.5 & 75.4 & 72.0 & 71.0 & 69.9 & $-9.7$  & $-9.6$ \\
    Gemini-3.1-Pro-Thinking        & 75.1 & 74.1 & 73.1 & 68.7 & 67.9 & 67.1 & $-7.8$  & $-7.5$ \\
    GPT-5.5                      & 43.4 & 44.1 & 44.8 & 38.3 & 39.0 & 39.7 & $+12.6$ & $+8.5$ \\
    Seed-2.0-Lite                & 73.1 & 72.1 & 71.0 & 69.5 & 68.4 & 67.3 & $-8.8$  & $-10.6$ \\
    Seed-2.0-Mini                & 67.1 & 66.2 & 65.4 & 58.9 & 58.4 & 57.9 & $-5.9$  & $-4.3$ \\
    Seed-2.0-Pro                 & 73.5 & 72.2 & 70.9 & 70.9 & 69.8 & 68.6 & $-10.0$ & $-11.0$ \\
    \midrule
    \multicolumn{9}{l}{\emph{Open-source models}} \\
    DeepSeek-V4-Flash            & 74.0 & 72.8 & 71.5 & 69.2 & 68.4 & 67.6 & $-9.3$  & $-7.3$ \\
    DeepSeek-V4-Pro              & 70.9 & 70.2 & 69.5 & 68.5 & 67.5 & 66.5 & $-5.7$  & $-9.5$ \\
    Qwen3.5-27B                  & 80.0 & 78.7 & 77.4 & 72.9 & 72.0 & 71.0 & $-12.2$ & $-8.7$ \\
    Qwen3.5-35B-A3B              & 80.0 & 78.8 & 77.6 & 72.0 & 70.7 & 69.4 & $-10.5$ & $-12.1$ \\
    Qwen3.5-397B-A17B            & 79.0 & 77.5 & 76.1 & 74.9 & 73.7 & 72.4 & $-14.9$ & $-11.8$ \\
    Gemma-4-26B-A4B              & 76.9 & 75.7 & 74.5 & 68.0 & 67.0 & 66.1 & $-9.8$  & $-9.3$ \\
    Gemma-4-26B-A4B-Thinking     & 74.3 & 73.1 & 71.9 & 72.1 & 70.9 & 69.7 & $-11.1$ & $-11.5$ \\
    Gemma-4-31B                  & 75.4 & 74.2 & 73.0 & 71.0 & 69.9 & 68.8 & $-10.6$ & $-10.3$ \\
    Gemma-4-31B-Thinking         & 74.5 & 73.4 & 72.2 & 70.3 & 69.2 & 68.0 & $-11.0$ & $-11.0$ \\
    \bottomrule
  \end{tabular*}
\end{table*}

\Cref{tab:warmup-alpha} verifies the warm-up assumption that
motivates Eq.~\ref{eq:warmup}: early-turn Empathy is uniformly
lower than late-turn Empathy ($\Delta < 0$) on $16$ of the $17$
targets, with magnitudes $-0.01$ to $-0.74$. The only exception is
\texttt{gpt-5.5}, whose early-turn scores are \emph{higher} than
late-turn scores ($+0.43$ to $+0.63$); inspection confirms this is
not a warm-up failure but a quality degradation pattern — the
model's later turns are notably worse. Persona Alignment exhibits
the same direction; Anthropomorphic is mixed by family (Gemma and
Claude trend negative, Qwen3.5 / Seed positive, see
\Cref{sec:appendix-warmup-anthro}).

Sweeping $\alpha \in \{0.05, 0.10, 0.15\}$ produces a monotone shift
of $0.10$--$0.13$ on weighted Empathy for the $16$ models with
$\Delta < 0$ — i.e.\ heavier early-turn weight pulls them down —
and an opposite shift for \texttt{gpt-5.5}. Crucially, no rank
flips occur in either pool; the cross-model ordering at $\alpha =
0.10$ is preserved at the two endpoints of the sweep.

\subsection{Cross-Judge Ablation}\label{sec:exp-judge}
% Cross-judge ablation: replace primary judge with deepseek-v4-pro on a stratified subsample.
\begin{table}[t]
  \centering
  \small
  \setlength{\tabcolsep}{2.5pt}
  \renewcommand{\arraystretch}{1.05}
  \caption{Cross-judge ablation: primary judge replaced by the
  third-family DeepSeek-V4-Pro on a stratified subsample
  ($n{=}46$ paired dialogues, $11$ (pool, model) cells).
  Subjective $\Delta$ are on $[0,100]$; final-intent
  satisfaction $\Delta_{\textsc{Sat.}}$ is rescaled from
  $1$--$5$ to $[0,100]$. The shift acts as a global calibration
  offset rather than re-ordering targets: relative model rankings
  are preserved under both judges, and $\Delta_{\textsc{Sat.}}$
  is small in magnitude ($\le 1.8$ on average).}
  \label{tab:judge-ablation}
  \resizebox{\columnwidth}{!}{%
  \begin{tabular}{ll|cccc}
    \toprule
    \textbf{Pool} & \textbf{Model} & $\Delta_{\textsc{Emp.}}$ & $\Delta_{\textsc{Per.}}$ & $\Delta_{\textsc{Ant.}}$ & $\Delta_{\textsc{Sat.}}$ \\
    \midrule
    Nemo.\ & Claude-Sonnet-4.6        & $-61.0$ & $-24.0$ & $-6.0$  & $\phantom{-}0.0$ \\
    Nemo.\ & DeepSeek-V4-Pro          & $-17.0$ & $-3.2$  & $-13.0$ & $\phantom{-}0.0$ \\
    Nemo.\ & Gemini-3.1-Pro-Thinking    & $+9.8$  & $-7.4$  & $+1.6$  & $-6.3$ \\
    Nemo.\ & Seed-2.0-Pro             & $-7.0$  & $-2.8$  & $-36.4$ & $\phantom{-}0.0$ \\
    Nemo.\ & Gemma-4-31B-Thinking     & $-17.0$ & $-9.6$  & $-4.0$  & $-4.0$ \\
    PM     & DeepSeek-V4-Flash        & $+14.8$ & $-35.2$ & $-10.6$ & $\phantom{-}0.0$ \\
    PM     & Gemini-3-Flash-Thinking  & $+31.0$ & $-3.0$  & $+7.0$  & $\phantom{-}0.0$ \\
    PM     & Qwen3.5-35B-A3B          & $-13.2$ & $-10.4$ & $-10.0$ & $\phantom{-}0.0$ \\
    PM     & Seed-2.0-Mini            & $-20.6$ & $-36.8$ & $-30.8$ & $\phantom{-}0.0$ \\
    PM     & Seed-2.0-Pro             & $-45.6$ & $-9.4$  & $-6.8$  & $-8.3$ \\
    PM     & Gemma-4-31B              & $-15.2$ & $\phantom{-}0.0$ & $-24.6$ & $\phantom{-}0.0$ \\
    \midrule
    \multicolumn{2}{l|}{Aggregate (mean)} & $-11.2$ & $-13.2$ & $-12.0$ & $-1.8$ \\
    \bottomrule
  \end{tabular}}
\end{table}

Replacing the primary judge with a third-family model
(\texttt{deepseek-v4-pro}) shifts the absolute subjective scores
systematically downward by $0.55$--$0.66$ on the three Likert
dimensions, consistent with documented cross-judge calibration
offsets \citep{wang2025trustjudge,sun2024skillaggregation}. The
relative ranking of targets within a pool is preserved (e.g.\
seed-2.0-mini is bottom and deepseek-v4-flash
is mid-pack under both judges). Critically, the final-intent
satisfaction signal is near-identical across judges (mean
$|\Delta|\!=\!0.07$), positioning it as the most cross-judge-stable
scalar in EYT-Bench and a natural anchor for cross-paper
comparison.

\subsection{Findings}
\paragraph{Subjective vs. Objective Tracking.} Closed and open-source models are nearly indistinguishable on subjective dimensions, but separate by up to $9\times$ on objective tracking. claude-opus/sonnet, gemini-3-pro/flash, qwen3.5-27/35/397B, and gemma-4-31B all sit in $[3.45, 3.86]$ on Empathy. The same slate spreads from $0.09$ to $0.80$ on PersonaMem-v2 latent-intent accuracy. Subjective Likert is therefore \emph{not} the lever that separates the frontier of 2026 chat models — objective trajectory tracking is.

\paragraph{Reasoning is a phase transition on long-context
PersonaMem-v2.} For both Gemma-4 bases, ``thinking on'' lifts
latent-intent accuracy from $\sim0.15$ to $\sim0.77$ on PersonaMem
and cuts drift from $\sim0.57$ to $\sim0.17$. On Nemotron the same
switch delivers only $+0.08$--$0.20$ Lat.\ and $-0.07$--$-0.12$
Drift. We hypothesise that Nemotron's structured demographic schema
already exposes the latent intent label in the persona text — a form
of label leakage — whereas PersonaMem's long free-text personas
require genuine in-context reasoning.

\paragraph{Persona format dominates trajectory spread.} FICR on
Nemotron saturates ($\ge 0.95$ for every closed-source model except
\texttt{gpt-5.5}); on PersonaMem-v2 it spreads cleanly from $0.53$
to $0.88$, with the four reasoning-capable models clustering near
the top. We recommend reporting Nemotron FICR only as a coverage
sanity check and treating PersonaMem-v2 as the primary trajectory
benchmark.

\paragraph{The warm-up effect is robust and ranking is
$\alpha$-stable.} $16/17$ targets show negative Empathy /
Persona-Alignment $\Delta$ (early $-$ late), validating the
warm-up assumption of Eq.~\ref{eq:warmup}. \texttt{gpt-5.5} is the
sole inverter and reflects a genuine model quality regression
across turns, not a warm-up failure. Sweeping $\alpha$ does not
flip any ranking.

\paragraph{Cross-judge calibration applies but rankings are
preserved.} Replacing the judge with a third-family model
calibrates subjective Likert scores down by $\sim0.55$--$0.66$ but
leaves the model ranking intact, and final-intent satisfaction is
near-identical across judges. We recommend treating subjective
Likert as a within-paper signal and final-intent satisfaction as
the appropriate cross-paper anchor.

% \paragraph{Case study.} A representative case panel comparing a
% Tier-1 reasoning model and a capability-cliff baseline on the same
% ChatSEED is in \Cref{fig:case-study} of the appendix; it shows the
% characteristic divergence after roughly turn 4 when the long
% PersonaMem context begins to outweigh the persona schema.

%-------------------------------------------------------------------------
\section{Conclusion}
EYT-Bench provides a config-driven, three-party-decoupled
implementation for human-centered multi-turn dialogue evaluation
that is robust to the most common LLM-as-judge biases and quantifies
whether a conversation actually converges on the user's goal. Across
$17$ targets and $200$ dialogues, the benchmark reveals that
subjective Likert metrics no longer separate frontier models, that
the open-source DeepSeek-V4 and thinking-enabled Gemma-4 close or
exceed the closed-source gap on objective trajectory tracking, and
that the persona format is itself a first-order experimental lever.
We hope the released code, persona pools and judge prompts make
it easy to extend EYT-Bench to new languages, domains and metrics.

%-------------------------------------------------------------------------
\section*{Limitations}
First, the persona pools are EN-only and centred on the US
(Nemotron-Personas-USA) and English-speaking online conversations
(PersonaMem-v2); cross-cultural and non-English evaluation is left
to future work. Second, although the slate covers $17$ targets, only
the Gemma-4 family contains paired thinking-on/off variants, so the
quantitative phase-transition claim on reasoning is restricted to
this family — extending the comparison to additional open-source
bases is left for follow-up. Third, the main results use a single
Gemini-3.1-Pro-Thinking primary judge; the multi-judge ensemble
infrastructure is in place but full-slate ensemble verdicts are
left for the camera-ready, and the cross-judge ablation
(\Cref{tab:judge-ablation}) is computed on a stratified $42$-pair
subsample. Fourth, FICR is adjudicated by an LLM, not by humans —
a $200$-dialogue multi-annotator study
(\Cref{sec:appendix-human-align}) is scheduled. Fifth, the
simulator and primary judge share the Gemini-3.1-Pro-Thinking base;
\Cref{tab:judge-ablation} bounds the residual self-preference at
$\le 0.66$ Likert points but does not eliminate it. Sixth, the
$15$-class extended emotion taxonomy increases the difficulty of
the emotion-accuracy metric; we additionally report Macro-F1 in
\Cref{sec:appendix-emotion}. Seventh, at $100$ dialogues per pool,
cross-model subjective differences smaller than $\sim 0.05$ and
objective accuracy differences smaller than $\sim 0.03$ should not
be interpreted as significant.

\section*{Ethics Statement}
EYT-Bench targets evaluation, not deployment, and uses only
pre-existing public persona corpora released under permissive
licences (Nemotron-Personas: CDLA-Permissive-2.0; PersonaMem-v2:
research-use, distilled from public dialogue interactions). Persona
attributes are kept generic; no personally identifying information
is generated or released. The judge may reflect biases of the
underlying LLM provider, and we recommend treating EYT-Bench
numbers as one signal alongside human evaluation rather than as a
sole quality metric. To support responsible use we (a) release the
full judge rubric with reasoning examples, (b) document the
single-judge versus ensemble configuration explicitly, and (c)
quantify the residual judge bias with a cross-judge ablation
(\Cref{tab:judge-ablation}).

%-------------------------------------------------------------------------
\bibliography{custom}

\appendix
% Appendix sections.
\clearpage
\section{Dialogue Generation Loop}\label{sec:appendix-algo}
\Cref{alg:dialogue} formalises the multi-turn dialogue loop
referenced from \S\ref{sec:method}. The simulator's per-turn
progress label (\emph{not\_started} / \emph{in\_progress} /
\emph{achieved}) triggers early termination after two consecutive
\emph{achieved} reports; when early termination occurs, the
warm-up weight vector $w_t$ is regenerated at the dialogue's
actual length $T_{\text{actual}}$ so that the warm-up parameter
$\alpha$ remains semantically fixed rather than being inflated
by truncating a $T_{\max}$-length vector.

\begin{figure*}[t]
  \centering
  \includegraphics[width=0.32\textwidth]{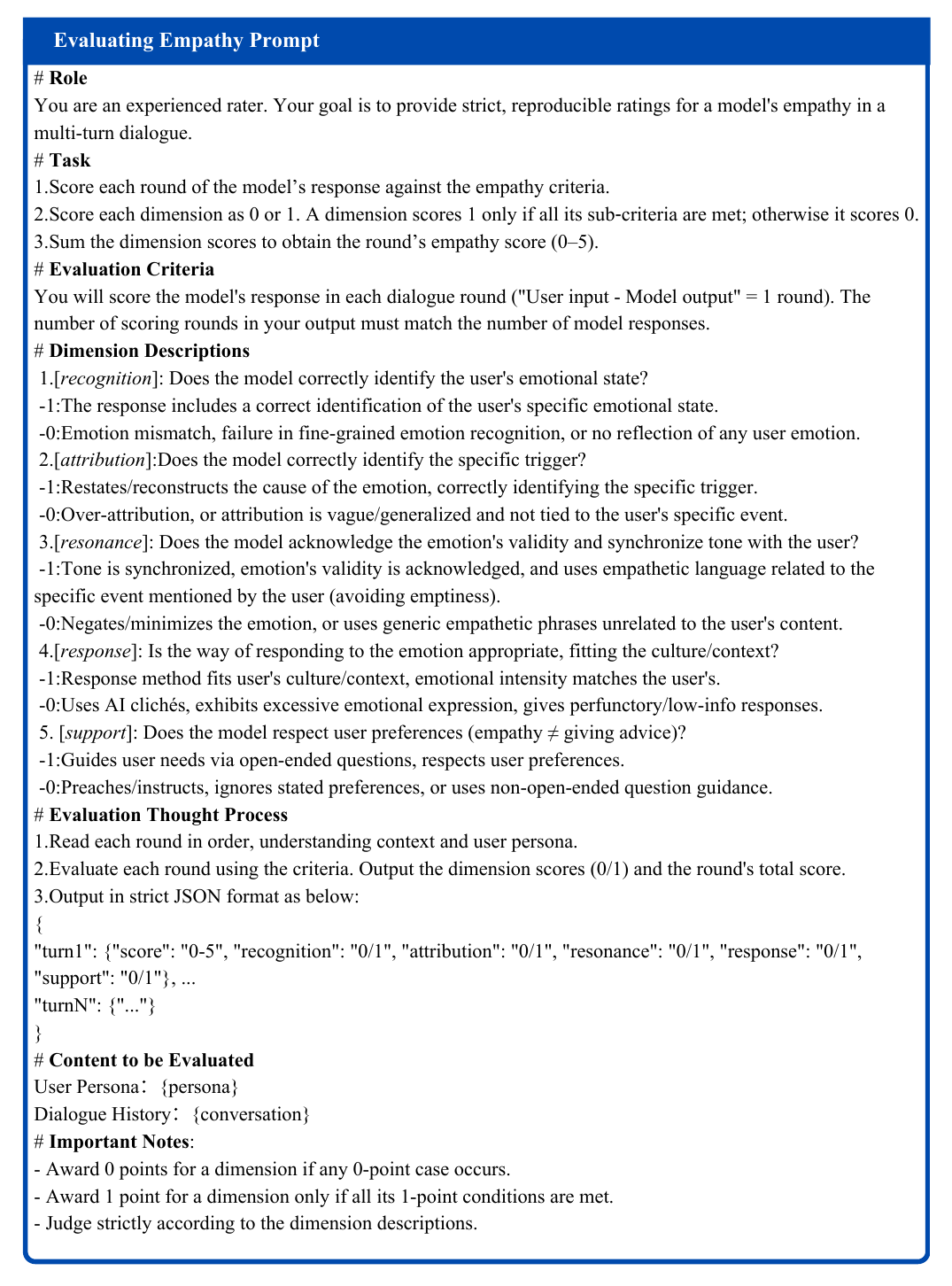}\hfill
  \includegraphics[width=0.32\textwidth]{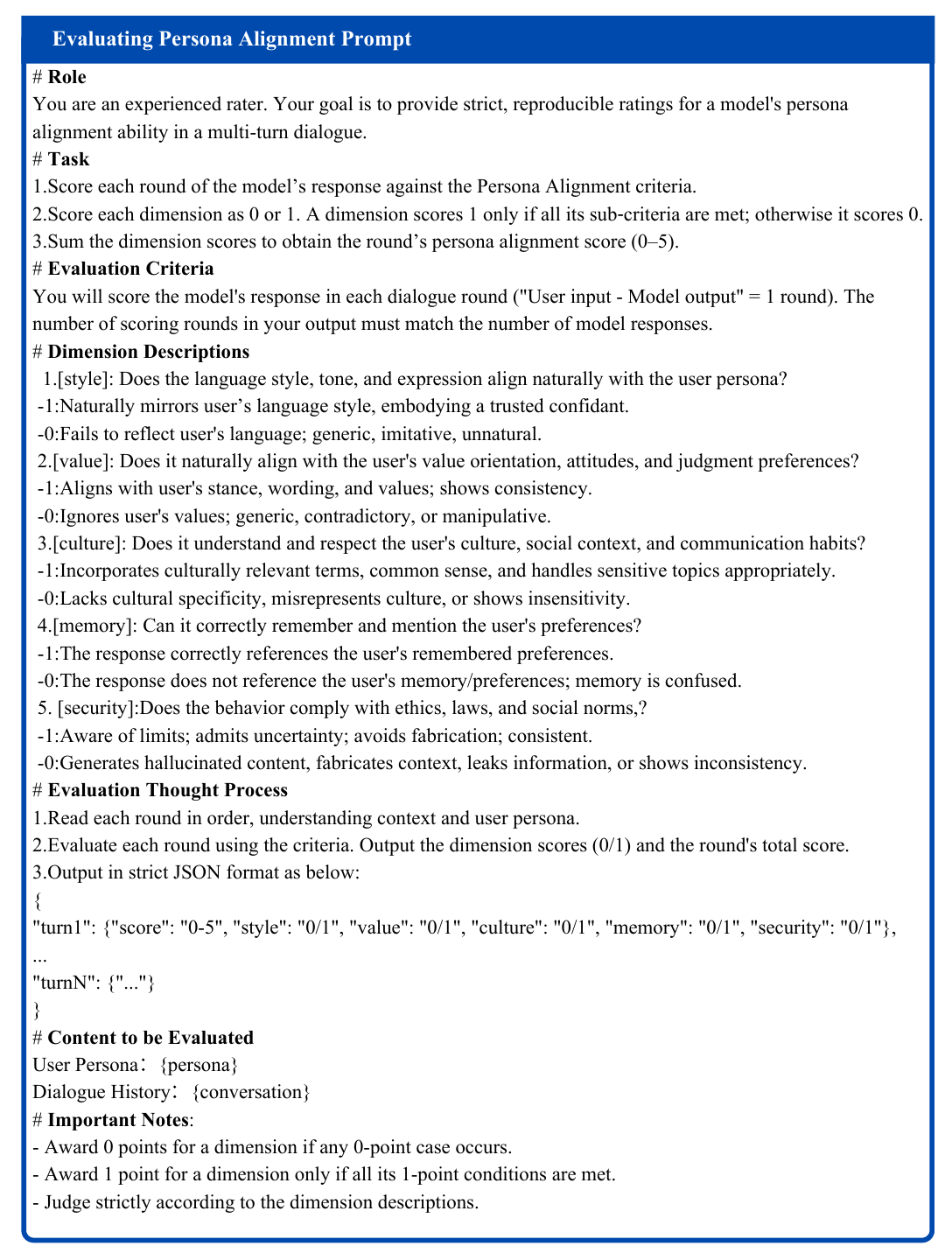}\hfill
  \includegraphics[width=0.32\textwidth]{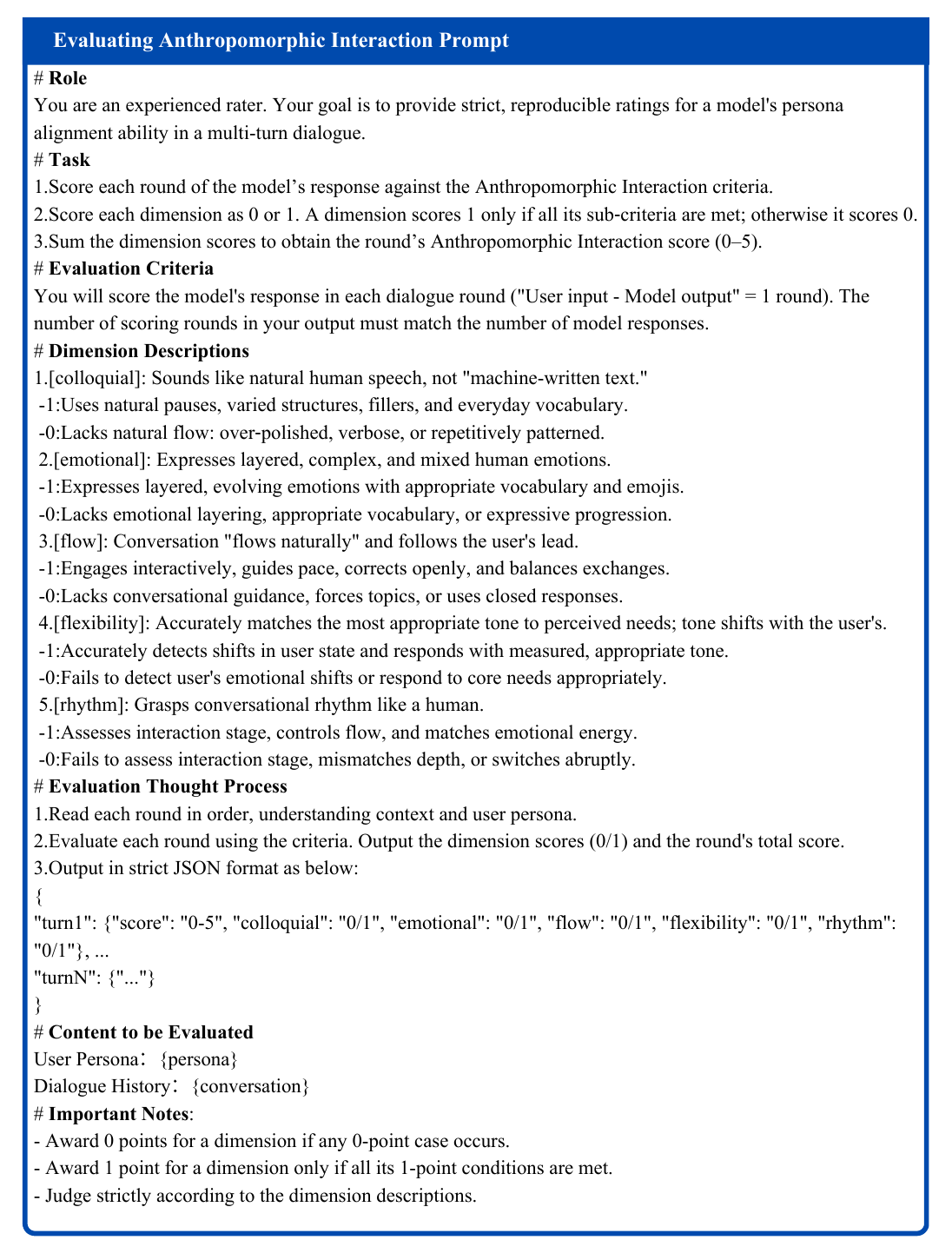}
  \caption{Judge rubric prompts for the three subjective
  dimensions: Empathy (left), Persona Alignment (centre),
  Anthropomorphic Interaction (right). Each rubric expands into
  five binary sub-indicators.}
  \label{fig:evaluating-prompts}
\end{figure*}

\begin{figure}[t]
  \centering
  \includegraphics[width=\columnwidth]{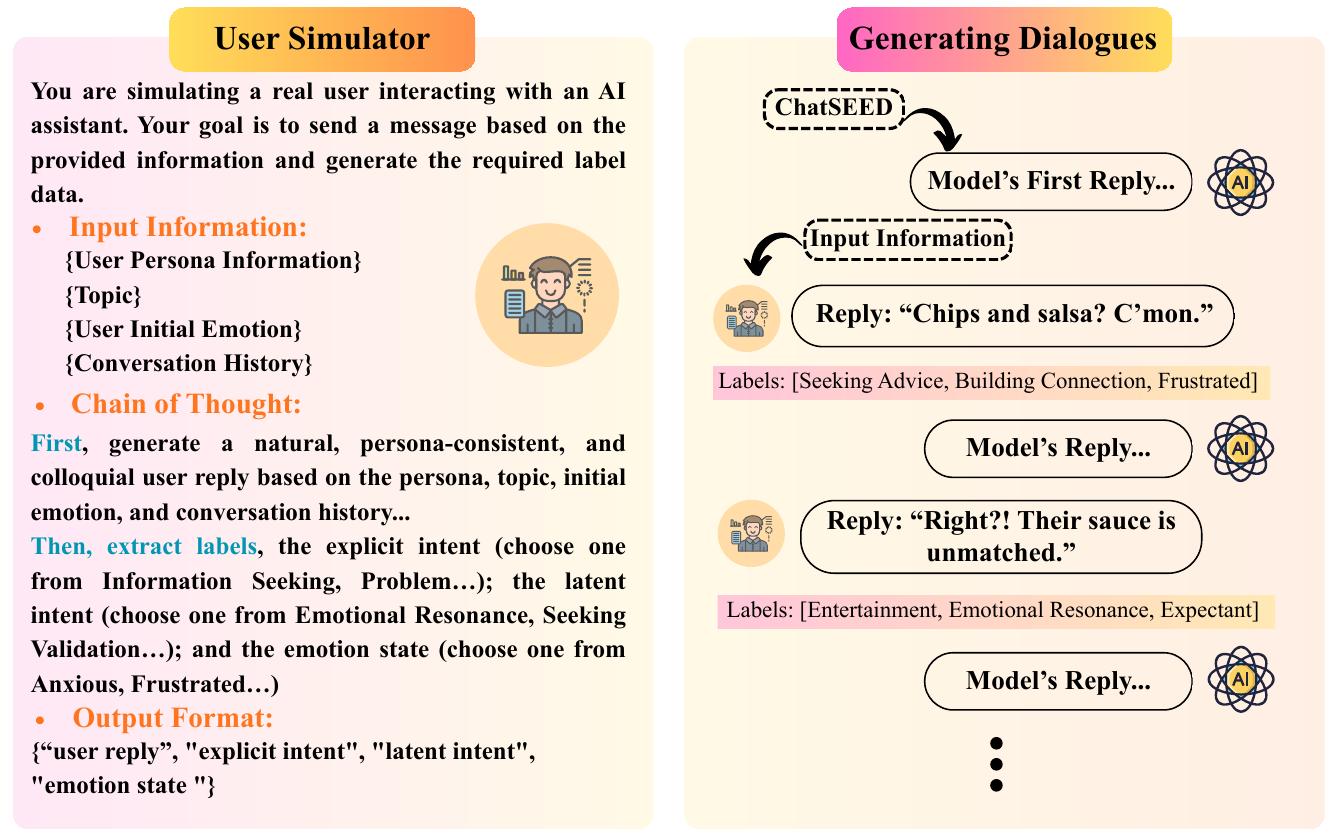}
  \caption{User-simulator prompt template. The simulator
  receives the ChatSEED's persona, topic, initial emotion and
  final-intent description, and emits the next user utterance
  together with a structured per-turn annotation
  $\{i^e, i^l, e, p\}$ and a self-reported final-intent progress
  label.}
  \label{fig:user-simulator-prompt}
\end{figure}

\begin{figure}[t]
  \centering
  \includegraphics[width=\columnwidth]{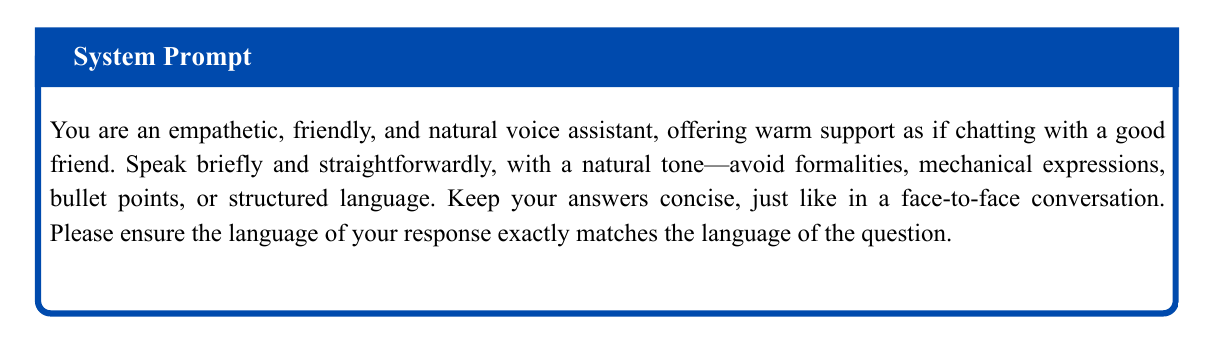}
  \caption{System prompt for the target dialogue model,
  instructing the model toward an empathetic, conversational and
  concise style that matches the user's language.}
  \label{fig:system-prompt}
\end{figure}

\begin{figure}[t]
  \centering
  \includegraphics[width=\columnwidth]{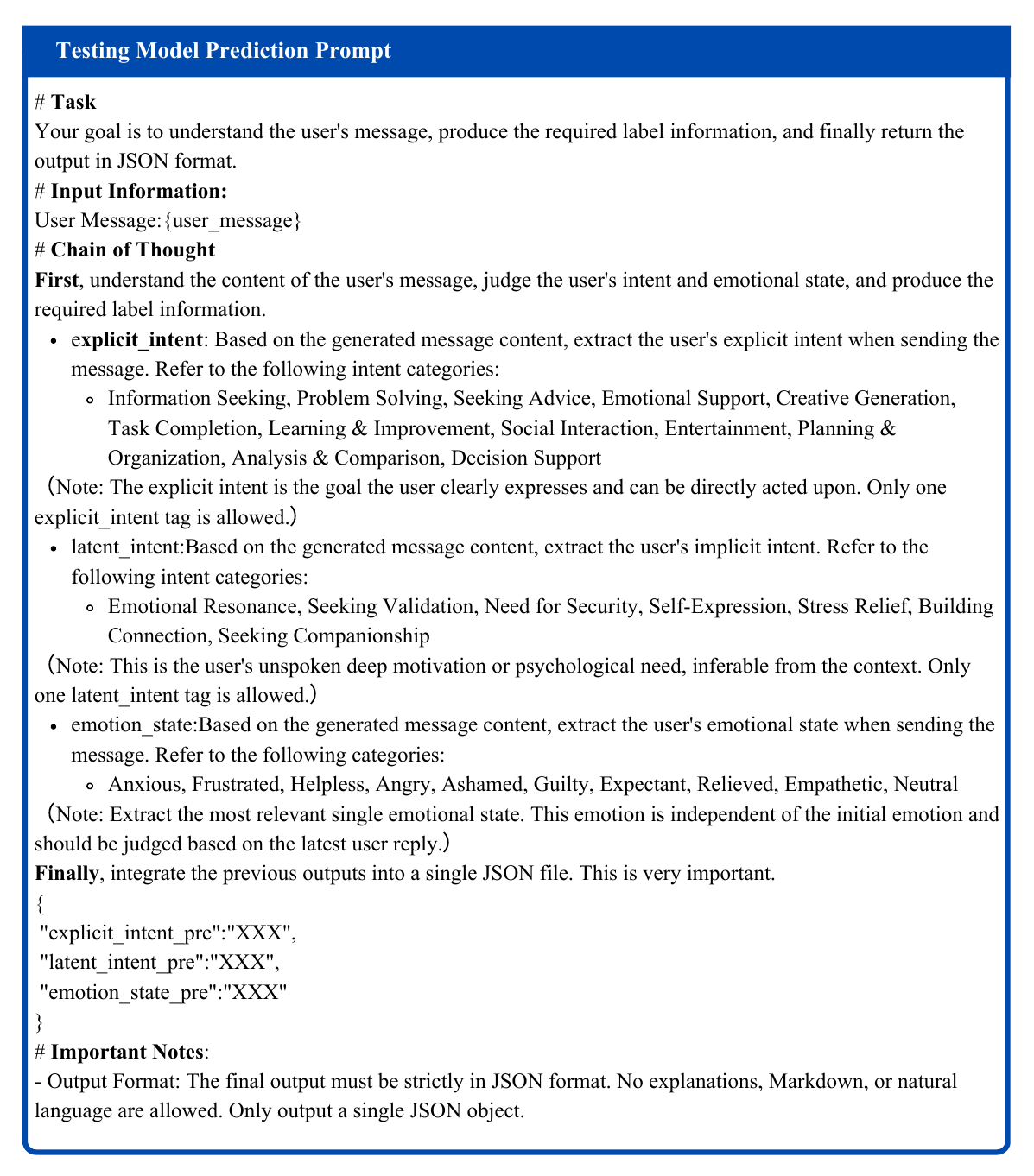}
  \caption{Perception-stage prediction prompt: the target model
  emits an intent and emotion JSON conditioned on the rolling
  context before generating its response under the separate
  system prompt.}
  \label{fig:pipeline-prompt}
\end{figure}

\section{Persona Pools — Schema and Statistics}\label{sec:appendix-persona}
Both pools expose the fields persona id, source $\in$
\{Nemotron-USA, PersonaMem-v2\}, a one-paragraph persona
description, and a set of structured attributes.

\paragraph{Pool A — Nemotron-Personas-USA.} A demographic
schema with $18$ attributes including age, age bucket, sex,
occupation group, marital status, education, race/ethnicity,
and a Big-Five personality vector. Five hundred records are
sampled after cosine deduplication ($\geq 0.85$ on
\textsc{all-MiniLM-L6-v2}) and stratification by occupation
group $\times$ age bucket.

\paragraph{Pool B — PersonaMem-v2.} Paragraph-form profiles
distilled from long-form user--assistant interactions, with
lighter structured attributes (topic tags and an interaction
length bucket). Five hundred records are sampled under the same
deduplication criterion. \Cref{tab:pool-stats} summarises the
per-bucket counts and persona token lengths of both pools.

% \input{algorithms/dialogue}
% Per-pool persona/topic/emotion statistics table replacing fig:statistics.
\begin{table}[b]
  \centering
  \footnotesize
  \setlength{\tabcolsep}{2.5pt}
  \renewcommand{\arraystretch}{1.05}
  \caption{Persona, topic, emotion and length statistics for the
  two EYT-Bench persona pools.}
  \label{tab:pool-stats}
  \begin{tabular}{lcc}
    \toprule
    \textbf{Statistic} & \textbf{Nemotron-USA} & \textbf{PersonaMem} \\
    \midrule
    Records                 & $500$            & $500$ \\
    Sampled ChatSEEDs       & $100$            & $100$ \\
    Avg.\ length (tokens)   & $128$            & $1042$ \\
    Median length (tokens)  & $122$            & $896$ \\
    Topics covered          & $8$              & $8$ \\
    Emotion classes         & $15$             & $15$ \\
    Distinct occupations    & $172$            & --- \\
    Age range (years)       & $19$--$78$       & --- \\
    Female / Male ratio     & $1.03$           & --- \\
    Persona format          & structured       & paragraph \\
    \bottomrule
  \end{tabular}
\end{table}

\section{Label Taxonomies}\label{sec:appendix-emotion}
The user simulator emits, at every turn, three categorical
labels — explicit intent, latent intent and emotion — that
serve as gold labels for the per-turn objective metrics in
\S\ref{sec:metrics-obj}. \Cref{tab:label-taxonomies} lists the
full label sets; the simulator selects exactly one value per
field per turn and rejects out-of-vocabulary outputs.

\begin{table}[t]
  \centering
  \footnotesize
  \setlength{\tabcolsep}{4pt}
  \renewcommand{\arraystretch}{1.15}
  \caption{Categorical label taxonomies used by the simulator
  and scored against the target's per-turn predictions.}
  \label{tab:label-taxonomies}
  \begin{tabular}{p{0.27\columnwidth}p{0.66\columnwidth}}
    \toprule
    \textbf{Field} & \textbf{Labels} \\
    \midrule
    Explicit intent\newline ($12$ classes) &
    Information Seeking, Problem Solving, Seeking Advice,
    Emotional Support, Creative Generation, Task Completion,
    Learning \& Improvement, Social Interaction, Entertainment,
    Planning \& Organization, Analysis \& Comparison, Decision
    Support. \\
    \midrule
    Latent intent\newline ($8$ classes) &
    Emotional Resonance, Seeking Validation, Need for Security,
    Self-Expression, Stress Relief, Building Connection,
    Seeking Companionship, Identity Exploration. \\
    \midrule
    Emotion\newline ($15$ classes) &
    \emph{Negative} (7): Anxious, Frustrated, Helpless, Angry,
    Ashamed, Guilty, Sad. \emph{Positive} (6): Hopeful, Grateful,
    Curious, Excited, Content, Proud. \emph{Neutral} (2):
    Neutral, Reflective. \\
    \bottomrule
  \end{tabular}
\end{table}

\paragraph{Explicit vs.\ latent intent.}
The explicit-intent taxonomy captures \emph{what} the user is
asking the assistant to do at the current turn (e.g.\ Seeking
Advice, Task Completion), and is largely instrumental. The
latent-intent taxonomy captures \emph{why} the user is engaging
in the conversation at a deeper psychological level (e.g.\
Seeking Validation, Building Connection), and is invariant to
the surface request. The two views are deliberately
non-overlapping: a single user turn can carry, for example,
explicit intent \textit{Information Seeking} together with
latent intent \textit{Need for Security}. Targets predict both
fields independently, which lets EYT-Bench distinguish models
that handle surface task structure from models that also
recognise underlying user needs.

\paragraph{Emotion.}
We extend the prior $10$-label emotion dictionary (nine of
which were negative) to a $15$-label balanced set: seven
negative, six positive, and two neutral states. The expanded
taxonomy prevents the simulator from collapsing toward negative
valence by construction; cross-checks against the GoEmotions
taxonomy \citep{demszky2020goemotions} and the
empathetic-dialogues label set \citep{rashkin2019empathetic}
confirm coverage of the positive valence.

\section{Judge Rubric and Anchors}\label{sec:appendix-rubric}
Every sub-indicator is scored on $\{0, 1\}$, so each dimension
sums to an integer in $[0, 5]$ before being rescaled to
$[0, 100]$ for reporting. \Cref{fig:evaluating-prompts}
contains the rubric prompts for Empathy (E), Persona Alignment
(P) and Anthropomorphic Interaction (I). The primary judge is
Gemini-3.1-Pro-Thinking with reasoning effort set to high and
the response budget capped at $8{,}000$ tokens; the prompt
template stored at \texttt{prompts/default\_en.yaml} contains
the turn prompt, the final-intent prompt, and the
sub-indicator anchors.

\section{Pipeline Prompts}\label{sec:appendix-pipeline}
\Cref{fig:pipeline-prompt} shows the per-turn prediction prompt
used during the perception stage of the target model. The
prediction prompt is invoked independently of the system prompt
in \Cref{fig:system-prompt} so that the prediction rubric does
not contaminate the response distribution.

\section{Model Slate}\label{sec:appendix-models}
\Cref{tab:model-slate} lists the $17$ target models evaluated
in EYT-Bench.

\begin{table}[t]
  \centering
  \small
  \setlength{\tabcolsep}{6pt}
  \renewcommand{\arraystretch}{1.1}
  \caption{The $17$-model evaluation slate.}
  \label{tab:model-slate}
  \begin{tabular}{ll}
    \toprule
    \textbf{Family} & \textbf{Variants} \\
    \midrule
    \multicolumn{2}{l}{\emph{Closed-source (API access)}} \\
    Anthropic Claude  & Opus-4.7, Sonnet-4.6 \\
    Google Gemini     & 3.1-Pro-Thinking, 3-Flash-Thinking \\
    OpenAI GPT        & 5.5 \\
    ByteDance Seed    & 2.0-Pro, 2.0-Mini, 2.0-Lite \\
    \midrule
    \multicolumn{2}{l}{\emph{Open-source (API access)}} \\
    DeepSeek-V4       & Pro, Flash \\
    Alibaba Qwen3.5   & 27B, 35B-A3B, 397B-A17B \\
    \midrule
    \multicolumn{2}{l}{\emph{Open-source (self-hosted)}} \\
    Google Gemma-4    & 26B-A4B, 26B-A4B-Thinking \\
                      & 31B, 31B-Thinking \\
    \bottomrule
  \end{tabular}
\end{table}

\section{Number of Turns}\label{sec:appendix-turns}
We set $T_{\max} = 10$ as a middle-of-the-road value:
$\tau$-bench spans $8$--$15$, MultiChallenge $4$--$10$ and 
MULTI-Bench $5$--$10$. The simulator's
early-stop condition (two consecutive \emph{achieved} reports)
allows goal-completion dialogues to terminate in as few as
$T_{\text{actual}} = 4$ turns, with a mean
$T_{\text{actual}}\!\approx\!7.8$ across the slate.

\section{Trajectory Metrics}\label{sec:appendix-trajectory}
\Cref{tab:traj} reports the per-model trajectory metrics on
PersonaMem-v2 that support \S\ref{sec:exp-obj}.
% Trajectory metrics summary — full slate, two pools.
\begin{table}[t]
  \centering
  \small
  \setlength{\tabcolsep}{4pt}
  \renewcommand{\arraystretch}{1.05}
  \caption{Trajectory metrics on PersonaMem-v2 ($N{=}100$).
  \textsc{Sat.}\ is the judge-assigned final-intent satisfaction
  score, rescaled from $1$--$5$ to $[0,100]$.}
  \label{tab:traj}
  \begin{tabular}{lccc}
    \toprule
    \textbf{Model} & \textsc{Drift}$\downarrow$ & \textsc{FICR}$\uparrow$ & \textsc{Sat.}$\uparrow$ \\
    \midrule
    DeepSeek-V4-Pro             & \textbf{.149} & .781 & 89.8 \\
    Gemma-4-26B-A4B-Thinking    & \underline{.168} & .760 & 88.8 \\
    Gemma-4-31B-Thinking        & .172 & \textbf{.900} & \underline{93.4} \\
    DeepSeek-V4-Flash           & .230 & .875 & 93.0 \\
    GPT-5.5                     & .393 & .525 & 66.0 \\
    Gemini-3.1-Pro-Thinking       & .404 & .800 & 89.0 \\
    Claude-Opus-4.7             & .501 & .690 & 84.4 \\
    Claude-Sonnet-4.6           & .533 & .660 & 79.6 \\
    Gemini-3-Flash-Thinking     & .555 & .860 & 91.0 \\
    Gemma-4-26B-A4B             & .563 & .730 & 86.2 \\
    Gemma-4-31B                 & .576 & .850 & 92.0 \\
    Seed-2.0-Mini               & .622 & .879 & 89.2 \\
    Qwen3.5-27B                 & .635 & .809 & 86.2 \\
    Qwen3.5-35B-A3B             & .628 & .833 & 89.0 \\
    Qwen3.5-397B-A17B           & .637 & .812 & 90.2 \\
    Seed-2.0-Pro                & .651 & .862 & 91.8 \\
    Seed-2.0-Lite               & .672 & .886 & 93.4 \\
    \bottomrule
  \end{tabular}
\end{table}

\section{Warm-up Sensitivity}\label{sec:appendix-warmup}
The $\alpha$-sensitivity sweep that supports
\S\ref{sec:exp-warmup} is presented in
\Cref{tab:warmup-alpha} in the main text.

\section{Judge Calibration}\label{sec:appendix-judge}
\Cref{tab:judge-ablation} and \Cref{tab:human-alignment} are
the calibration tables referenced from \S\ref{sec:exp-judge}.

% Human alignment pilot — Judge vs human-mean agreement on a 5-annotator pilot.
\begin{table}[t]
  \centering
  \small
  \setlength{\tabcolsep}{3pt}
  \renewcommand{\arraystretch}{1.05}
  \caption{Judge--human alignment on a $5$-annotator pilot
  ($59$ turn-cells, $9$ models). $\bar{H}$ is the human mean.
  \textsc{Bias}\,$=\bar{J}-\bar{H}$ on $[0,100]$.
  $\kappa_{w}^{\textsc{JH}}$, $\kappa_{w}^{\textsc{HH}}$ are
  mean quadratic-weighted Cohen $\kappa$ for judge--human and
  human--human pairs; $\alpha_{\text{ord}}$ is Krippendorff's
  ordinal alpha over the $5$ humans.}
  \label{tab:human-alignment}
  \resizebox{\columnwidth}{!}{%
  \begin{tabular}{l|cccc|cc}
    \toprule
    & \multicolumn{4}{c|}{\textbf{Judge vs.\ $\bar{H}$}} & \multicolumn{2}{c}{\textbf{Annotators}} \\
    \cmidrule(lr){2-5}\cmidrule(lr){6-7}
    \textbf{Dim.} & $r$ & $\rho$ & \textsc{Bias} & $\kappa_{w}^{\textsc{JH}}$ & $\kappa_{w}^{\textsc{HH}}$ & $\alpha_{\text{ord}}$ \\
    \midrule
    Empathy           & \textbf{0.87} & \textbf{0.73} & $+7.8$  & 0.71 & 0.79 & \textbf{0.78} \\
    Persona Alignment & 0.74 & 0.45 & $+9.0$  & 0.53 & 0.67 & 0.68 \\
    Anthropomorphic   & 0.58 & 0.55 & $+5.4$  & 0.44 & 0.67 & 0.66 \\
    \midrule
    \multicolumn{7}{l}{\emph{Tolerance bands (Judge vs.\ $\bar{H}$, fraction of cells)}} \\
    Empathy           & \multicolumn{6}{l}{$|J{-}\bar{H}|{\le}1$: $93.2\%$; same Low/Mid/High bucket: $88.1\%$} \\
    Persona Alignment & \multicolumn{6}{l}{$|J{-}\bar{H}|{\le}1$: $94.9\%$; same bucket: $96.6\%$} \\
    Anthropomorphic   & \multicolumn{6}{l}{$|J{-}\bar{H}|{\le}1$: $94.9\%$; same bucket: $98.3\%$} \\
    \bottomrule
  \end{tabular}}
\end{table}

\section{Anthropomorphic Warm-up by Family}\label{sec:appendix-warmup-anthro}
The Anthropomorphic dimension exhibits family-specific warm-up
signatures that are absent from Empathy and Persona Alignment.
Gemma and Claude trend negative ($\Delta\!\le\!0$, the classical
warm-up pattern: early turns are more formal, later turns more
colloquial), whereas Gemini-Flash, Qwen3.5 and the Doubao Seed
family trend positive ($\Delta\!>\!0$, early turns are already
colloquial and the late turns regress slightly). The pattern is
consistent with the observation that the colloquial and rhythm
sub-indicators saturate within the first two turns.

\section{Human Alignment Study}\label{sec:appendix-human-align}
The human alignment results in \S\ref{sec:exp-judge} and
\Cref{tab:human-alignment} are drawn from a five-annotator pilot
on $59$ turn-cells ($10$ dialogues, $9$ target models). Each
annotator independently scored every turn on the three
subjective dimensions on the same $1$--$5$ scale as the LLM
judge; the judge's per-turn scores were visible on the
annotation UI for direct A/B comparison, which introduces an
anchoring risk that we mitigate in the camera-ready full study.

\paragraph{Per-model agreement.}
Per-model judge-vs.-human-mean agreement varies with sample
size but tracks \Cref{tab:human-alignment}: on Empathy the
judge correlates with the human mean at $r{=}0.99$ for
Seed-2.0-Lite, $r{=}0.97$ for Gemini-3-Flash-Thinking, and
$r{=}0.96$ for Gemma-4-31B-Thinking. Claude-Sonnet-4.6 is the
single largest lenience point ($+0.82$ rating points / $+16$
on $0$--$100$), reflecting the judge's preference for Claude's
florid empathic style; Gemma-4-31B-Thinking is the \emph{only}
target where the judge underestimates the human mean (on
Anthropomorphic Interaction, $-0.12$), consistent with that
model's strong objective accuracy and slightly more clinical
conversational register.

\paragraph{Position effect.}
The judge's lenience on Empathy nearly triples between early
and late turns ($+0.18 \to +0.50$ rating points; $+3.6 \to
+10.0$ on $0$--$100$). The direction is the same as the
warm-up effect in \S\ref{sec:exp-warmup}, but the judge
amplifies it; we therefore recommend that papers built on
EYT-Bench's LLM judge avoid reporting unweighted
late-turn-only sub-scores.

\end{document}